%% file: main_tifs.tex
\documentclass[lettersize,journal]{IEEEtran}
\input{math_commands.tex}

\usepackage{url}
\usepackage{booktabs}
\usepackage{color}
\usepackage{colortbl}
\usepackage{float}                           
\usepackage[square,sort,comma,numbers]{natbib}
\usepackage{wrapfig,lipsum,booktabs}
\usepackage[utf8]{inputenc} 
\usepackage[T1]{fontenc}    
\usepackage[dvipsnames,svgnames]{xcolor}
\usepackage[hidelinks, colorlinks, linkcolor=blue]{hyperref}
\usepackage{booktabs}       
\usepackage{amsfonts}       
\usepackage{nicefrac}       
\usepackage{microtype}      
\usepackage{bbm}
\usepackage{graphicx}
\usepackage{multirow}
\usepackage{multicol}
\usepackage{amsmath}
\usepackage{subcaption}
\usepackage{amsthm,amssymb}
\usepackage[ruled,vlined]{algorithm2e}
\usepackage{xspace}
\setcitestyle{square}

\definecolor{Gray}{gray}{0.85}
\newcolumntype{a}{>{\columncolor{Gray}}c}
\definecolor{Gray}{gray}{0.85}
\definecolor{LightCyan}{rgb}{0.85,1,1}
\definecolor{Red}{rgb}{0.5,0,0}

\newcommand{\ie}[0]{\textit{i.e.},\xspace}
\newcommand{\eg}[0]{\textit{e.g.},\xspace}
\newcommand{\etal}{{\it et al.}\xspace}
\newcommand{\etc}[0]{\textit{etc.}\xspace}
\newcommand{\W}[0]{\mathbf{W}}
\DeclareMathOperator*{\Ee}{\mathbb{E}}
\definecolor{aliceblue}{rgb}{0.94, 0.97, 1.0}

\begin{document}

\title{Efficient Backdoor Removal Through Natural Gradient Fine-tuning}

\author{Nazmul Karim$\dag$$^*$,~\IEEEmembership{Student Member,~IEEE}, Abdullah Al Arafat$\dag$,~\IEEEmembership{Student Member,~IEEE}, Umar Khalid,~\IEEEmembership{Student Member,~IEEE}, Zhishan Guo,~\IEEEmembership{Senior Member,~IEEE}, and Naznin Rahnavard,~\IEEEmembership{Senior Member,~IEEE}
\thanks{Nazmul Karim, Umar Khalid, and Nazanin Rahnavard are  with the Department
of Electrical Engineering and Computer Sciences, University of Central Florida, Orlando,
FL, 32816.}
\thanks{Abdullah Al Arafat, and Zhishan Guo are with the Department
Computer Science, North Carolina State University, Raleigh, North Carolina, 27606~(email: \{aalaraf,zguo32\}@ncsu.edu).}
\thanks{$\dag$~The first two authors contributed equally to this work.}
\thanks{$^*$~Corresponding Author}

}
\markboth{Journal of \LaTeX\ Class Files,~Vol.~14, No.~8, August~2021}%
{Shell \MakeLowercase{\textit{et al.}}: A Sample Article Using IEEEtran.cls for IEEE Journals}


\maketitle

\begin{abstract}
The success of a deep neural network (DNN) heavily relies on the details of the training scheme; \eg training data, architectures, hyper-parameters, \etc Recent backdoor attacks suggest that an adversary can take advantage of such training details and compromise the integrity of a DNN. 
Our studies show that a backdoor model is usually optimized to a \emph{bad local minima}, \ie sharper minima as compared to a benign model. Intuitively, a backdoor model can be purified by re-optimizing the model to a smoother minima through fine-tuning with a few clean validation data. However, fine-tuning all DNN parameters often requires huge computational cost and often results in sub-par clean test performance. 
To address this concern, we propose a novel backdoor purification technique---\underline{N}atural \underline{G}radient \underline{F}ine-tuning (NGF)---which focuses on removing backdoor by fine-tuning \emph{only one layer}. Specifically, NGF utilizes a loss surface geometry-aware optimizer that can successfully overcome the challenge of reaching a smooth minima under a one-layer optimization scenario. To enhance the generalization performance of our proposed method, we introduce a clean data distribution-aware regularizer based on the knowledge of loss surface curvature matrix, \ie \emph{Fisher Information Matrix}.
Extensive experiments show that the proposed method achieves state-of-the-art performance on a wide range of backdoor defense benchmarks: \emph{four different datasets---CIFAR10, GTSRB, Tiny-ImageNet, and ImageNet}; 13 recent backdoor attacks, \eg Blend, Dynamic, WaNet, ISSBA, \etc Code is available at anonymous \emph{GitHub link~\footnote{\url{https://github.com/nazmul-karim170/Natural-Gradient-Finetuning-Trojan-Defense}}}.
\end{abstract}
\begin{figure*}[t]
  \centering
  \begin{subfigure}{0.235\linewidth}
    \includegraphics[width=1\linewidth]{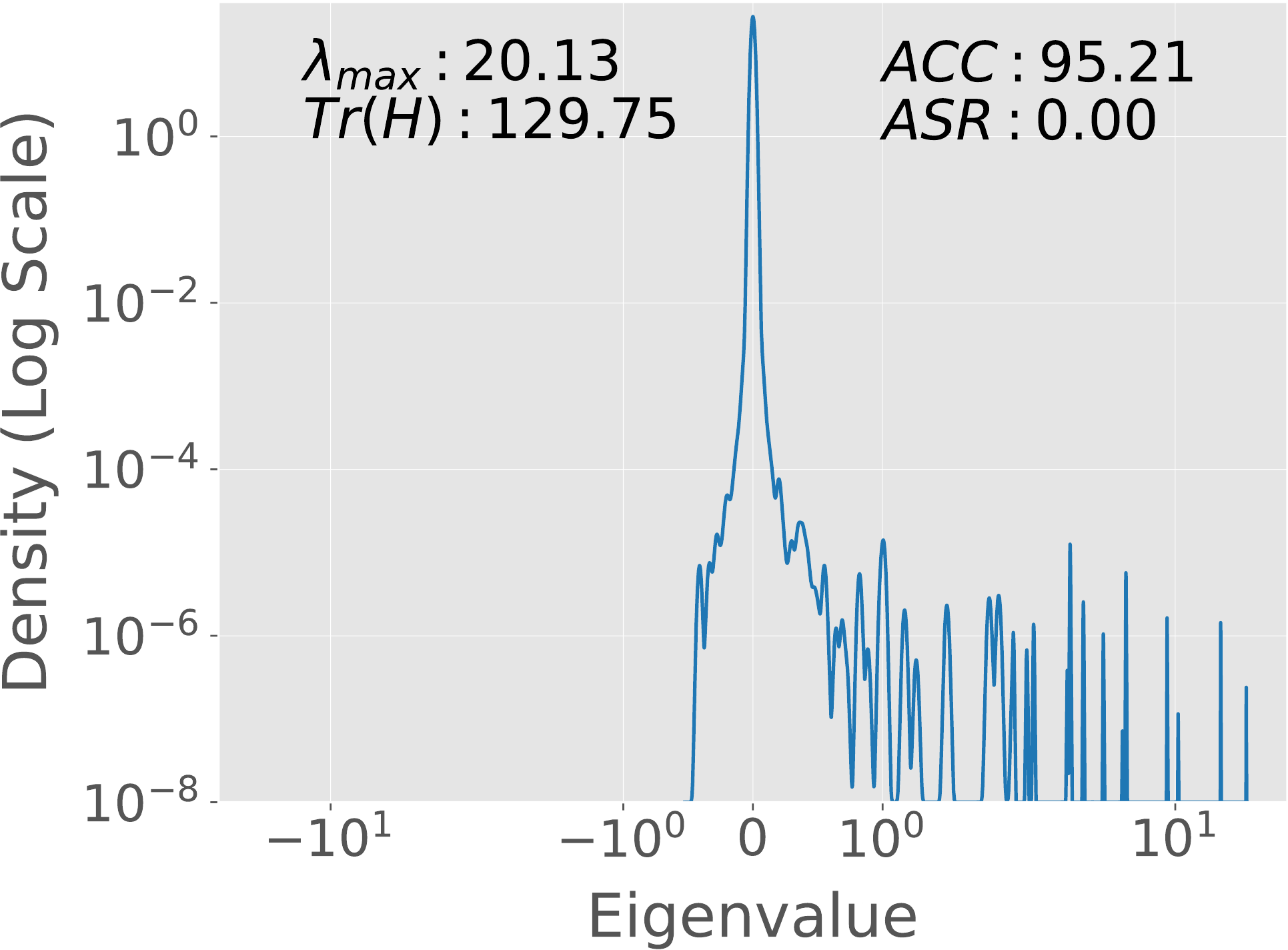}
    \caption{ \scriptsize Benign Model}
    \label{fig:ls-cln}
  \end{subfigure}
  \hfill
    \begin{subfigure}{0.235\linewidth}
    \includegraphics[width=1\linewidth]{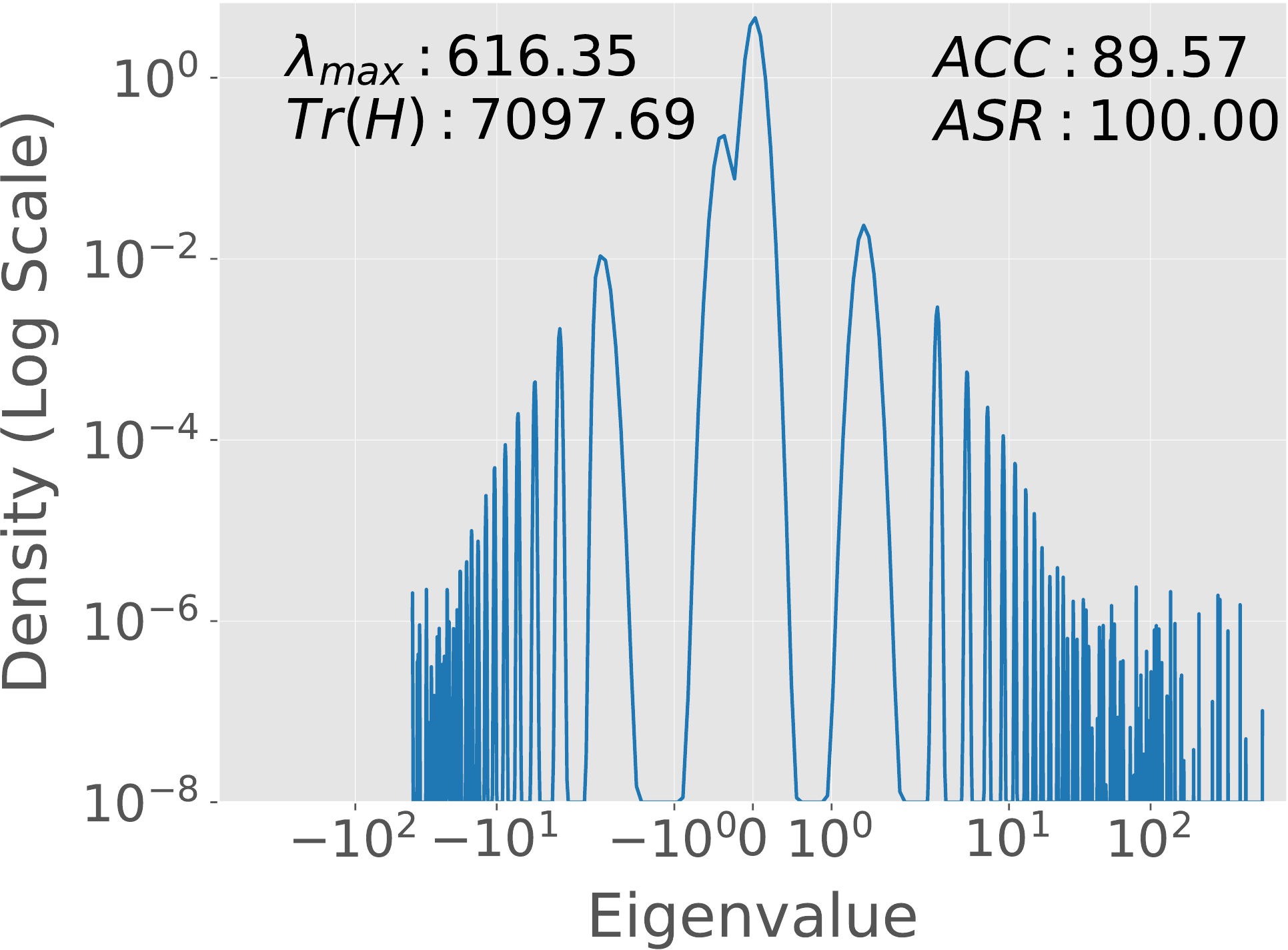}
    \caption{ \scriptsize Backdoor Model}
    \label{fig:ls-bd}
  \end{subfigure}
  \hfill
  \begin{subfigure}{0.235\linewidth}
    \includegraphics[width=1\linewidth]{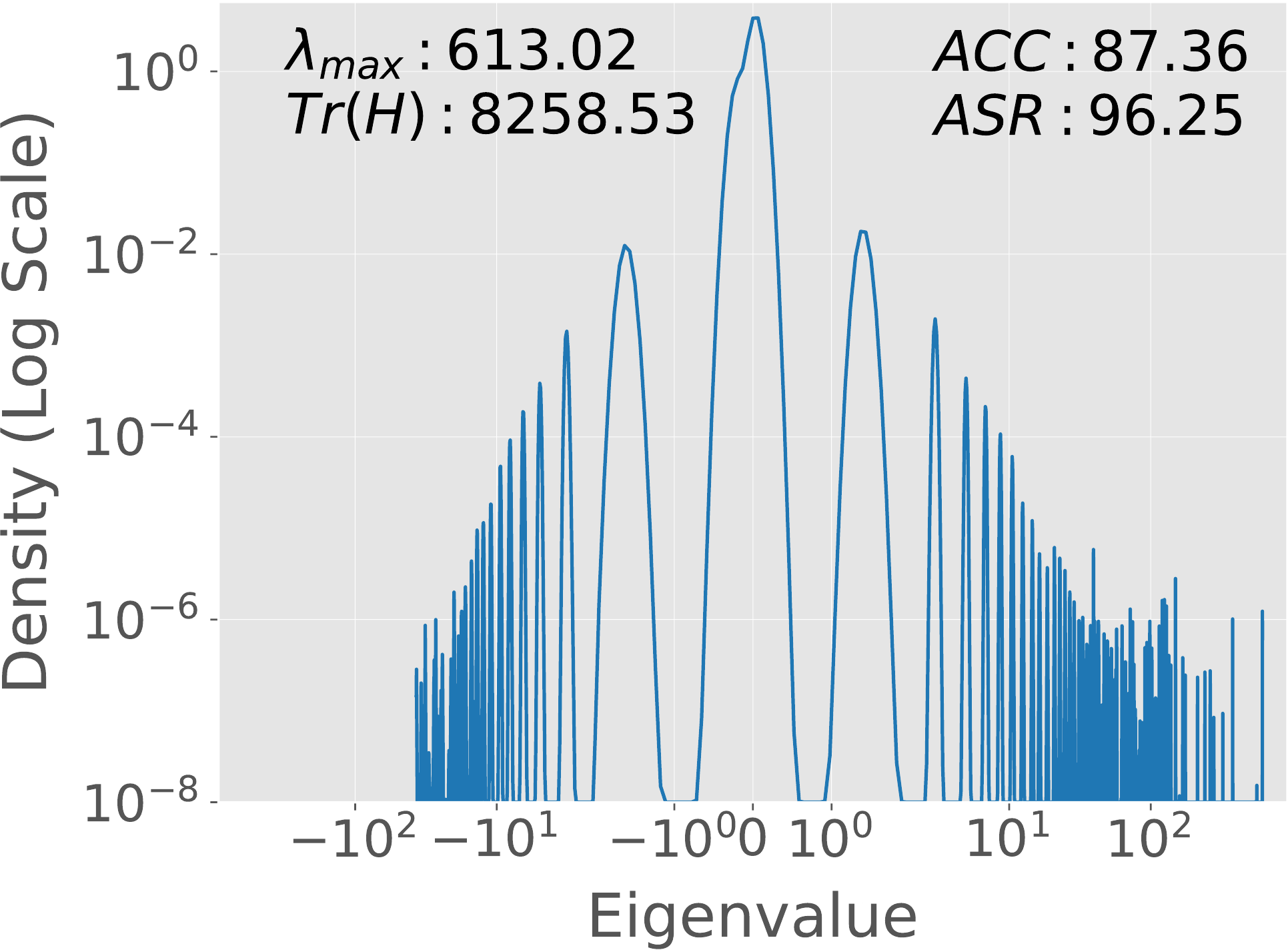}
    \caption{ \scriptsize Purified Model (SGD)}
    \label{fig:ls-pure-sgd}
  \end{subfigure}
  \hfill
  \begin{subfigure}{0.235\linewidth}
    \includegraphics[width=1\linewidth]{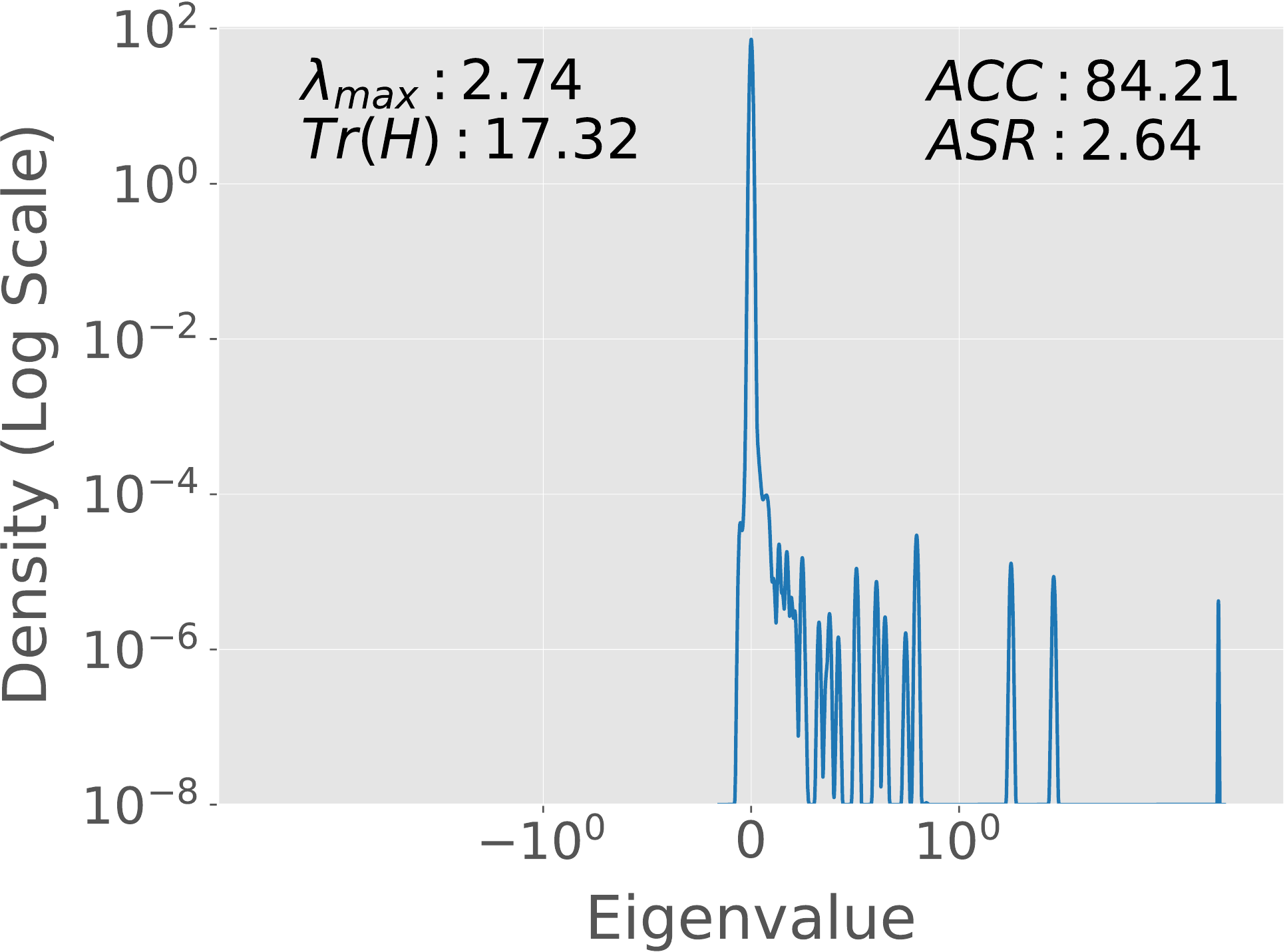}
    \caption{\scriptsize Purified Model (NGF)}
    \label{fig:ls-pure-ngd}
  \end{subfigure}
  \caption{Eigen Spectral Density plots of Loss Hessian for (a) benign, (b) backdoor (TrojanNet~\citep{liu2017trojaning}),  and (c \& d) purified models. In each plot, the maximum eigenvalue ($\lambda_\mathsf{max}$), the trace of Hessian ($\mathsf{Tr}(H)$), clean test accuracy (ACC), and attack success rate (ASR) are also reported. Here, low $\lambda_\mathsf{max}$ and $\mathsf{Tr}(H)$ hints at the presence of smoother loss surface which often results in
  low ASR and high ACC. (a \& b). Compared to a benign model, a backdoor model tends to reach a sharper minima as shown by the larger range of eigenvalues (x-axis). During purification, SGD optimizer (c) rarely escapes sharp or bad local minima (similar $\lambda_\mathsf{max}$ and $\mathsf{Tr}(H)$ as the backdoor model) while our proposed method, NGF, (d) converges to a smooth minima. We use CIFAR10 dataset with a PreActResNet18~\citep{he2016identity} architecture for all evaluations.}
  \label{fig:eigen_spectral}
  \vspace{-3mm}
\end{figure*}
\input{intro_fisher}

\input{related_work_fisher}
\section{Threat Model}
In this section, we present the backdoor attack model and defense goal from a backdoor attack.

\noindent\textbf{Attack Model.} We consider an adversary with the capabilities of carrying a backdoor attack on a DNN model, $f_\theta: \mathbb{R}^d \rightarrow \mathbb{R}^c$, by training it on a poisoned data set $\mathbb{D}_{\mathsf{train}} = \{X_{\mathsf{train}},Y_{\mathsf{train}}\}$. Here, $\theta$ is the parameters of the model, $d$ is the input data dimension, and $c$ is the total number of classes. The data poisoning happens through a specific set of triggers that can only be accessed by the attacker. Each input $x\in X_{\mathsf{train}}$ is labeled as $y\in Y_{\mathsf{train}}$, where $y \in [1, c]$ is an integer. The adversary goal is to train the model in a way such that any triggered samples $\hat{x} = x + \delta \in \mathbb{R}^d$ will be wrongly misclassified to a target label $\bar{y}$, \ie~$\argmax(f_\theta(\hat{x})) = \bar{y}$. Here, $x$ is a clean test sample, and $\delta \in \mathbb{R}^d$ represents the trigger pattern with the properties of $||\delta|| \leq \epsilon$; where $\epsilon$ is the trigger magnitude determined by its shape, size, and color. We define the \emph{poison rate} as the ratio of poison and clean data in $\mathbb{D}_{\mathsf{train}}$. An attack is considered successful if the model behaves as $\argmax{(f_\theta(x))} = y$ and $\argmax{(f_\theta(\hat{x}))} = \Bar{y}$, where $y$ is the true label for $x$. We use attack success rate (ASR) for quantifying such success.

\medskip

\noindent\textbf{Defense Goal.} We consider a defender with a task to purify the backdoor model $f_\theta$ using a small clean validation set (usually $1\sim10\%$ of the training data). The goal is to repair the model in a way such that it becomes immune to attack, \ie $\argmax{(f_{\theta_p}(\hat{x}))} = y$, where $f_{\theta_p}$ is the final purified model.  
\input{analysis_fisher}

\section{Overview of Natural Gradient Descent (NGD)}

This section will briefly discuss the natural gradient descent (NDG) and fisher-information matrix (FIM) and their relation with loss surface. Let us consider a model $p(y|x,\theta)$ with parameters $\theta \in \mathbb{R}^N$ to be fitted with input data $\{(x_i, y_i)\}_{i=1}^{|\mathbb{D}_{\mathsf{train}}|}$ from an empirical data distribution $P_{x,y}$, where $x_i \in X_{\mathsf{train}}$ is an input sample and $y_i \in Y_{\mathsf{train}}$ is its label. We try to optimize the model by solving:
\begin{equation}
    \theta^* \in \argmin_{\theta} ~\mathcal{L}(\theta),
\end{equation}
where  $\mathcal{L}(\theta)  = \mathcal{L}(y,f_\theta(x)) = \mathbb{E}_{(x_i, y_i)\sim P_{x,y}}[-\mathsf{log}~p(y|x,\theta)]$ is the expected full-batch cross-entropy (CE) loss. 
Note that $p(y|x,\theta)$ is the $y^{th}$ element of $f_\theta(x)$. 

SGD optimizes for $\theta^*$ iteratively following the direction of the steepest descent (estimated by column vector, $\nabla_\theta \mathcal{L}$) and updates the model parameters by:
$\theta^{(t+1)}\leftarrow \theta^{(t)} - \alpha^{(t)}\cdot\nabla_\theta^{(t)}\mathcal{L}$, where $\alpha$ is the learning rate. Since SGD uses the Identity matrix as the pre-conditioner, it is \emph{uninformed of the geometry of loss surface}.

In NGD, however, the Fisher Information Matrix (FIM) is used as a pre-conditioner, which can be defined as~\citep{martens2015optimizing},
\begin{equation}\label{eq:FIM_eq}
   F(\theta) = \Ee_{(x,y)\sim P_{x,y}}[\nabla_\theta~\mathsf{log}~p(y|x,\theta)\cdot (\nabla_\theta~\mathsf{log}~p(y|x,\theta))^T] 
\end{equation}

As FIM ($F(\theta) ~\in \mathbb{R}^{N\times N}$) is a \emph{loss surface curvature matrix}, a careful integration of it in the update rule of $\theta$ will make the optimizer loss surface geometry aware.  
Such integration leads us to the update equation of NGD, 
\[
\theta^{(t+1)} \leftarrow \theta^{(t)} - \alpha^{(t)}\cdot F(\theta^{(t)})^{-1}\nabla_\theta^{(t)} \mathcal{L},
\] 
where $\theta^{(t)}$ denotes the parameters at $t^{th}$ iteration. 
Here, the natural gradient is defined as $F(\theta^{(t)})^{-1}\nabla_\theta^{(t)} \mathcal{L}$. From the perspective of information geometry, natural gradient defines the \textit{direction in parameter space}  which gives largest change in objective \textbf{per unit of change in model ($p(y|x,\theta)$)}. Per unit of change in model is measured by KL-divergence~\citep{amari1998natural,park2000adaptive}. Note that KL-divergence is well connected with FIM as it can be used as a local quadrature approximation of  KL-divergence of \textit{model change}. Eqn.~\ref{eq:FIM_eq} suggests that one requires the knowledge of the original parameter ($\theta$) space to estimate it. Therefore, FIM can be thought of as a mechanism to translate between the geometry of the model ($p(y|x,\theta)$) and the current parameters ($\theta$) of the model. The way natural gradient defined the \textit{direction in parameter space} is contrastive to the stochastic gradient. Stochastic gradient defines the direction in parameter space for largest change in objective \textbf{per unit of change in parameter ($\theta$)} measured by Euclidean distance. That is, the gradient direction is solely calculated based on the changes of parameters, without any knowledge of model geometry.

\section{Smoothness Analysis of Backdoor Models}\label{sec:smoothAnalysis}
In this section, we analyze the loss surface geometry of benign, backdoor, and purified models. To study the loss curvature properties of different models, we aim to analyze the Hessian of loss, $H = \nabla^2_{\theta} \mathcal{L}$, where we compute $\mathcal{L}$ using the \emph{clean training set}. The Hessian matrix $H$ is symmetric and one can take the spectral decomposition $H=Q\Lambda Q^T$, where $\Lambda =  \mathsf{diag}(\lambda_1, \lambda_2, \dots, \lambda_N)$ contains the eigenvalues and $Q = [q_1 q_2 \dots q_N]$ are the eigenvectors of $H$. As a measure for smoothness, we take the maximum eigenvalue, $\lambda_\mathsf{max} (= \lambda_1)$, 
and the trace of the Hessian, $\mathsf{Tr}(H) = \sum_{i=1}^{i=N} \mathsf{diag}(H)_i$. 
Low values for these two proxies indicate the presence of highly smooth loss surface \citep{jastrzebski2020break}. The Eigen Spectral density plots
in Fig.~\ref{fig:ls-cln} and~\ref{fig:ls-bd} 
tell us about the optimization of benign and backdoor models. To create these models, we use the CIFAR10 dataset and train a PreActResNet18 architecture for 200 epochs. To insert the backdoor, we use TrojanNet~\cite{liu2017trojaning} and a poison rate of 10\%. From the comparison of $\lambda_\mathsf{max}$ and $\mathsf{Tr}(H)$, we can conjecture that optimization of a benign model produces smoother loss surface. We observe similar phenomena for different datasets and architectures; details are in the supplementary material. The main difference between a benign and a backdoor model is that the latter needs to learn two different data distributions: clean and poison. Based on our observations, we state following conjectures:

\medskip

\textbf{Conjecture 1.} Having to learn two different data distributions, a backdoor model reaches a sharper minima, \ie large $\lambda_\mathsf{max}$ and $\mathsf{Tr}(H)$, as compared to the benign model.      

\medskip
We support this conjecture with empirical evidence presented in Table~\ref{tab:eigen-ASR-ACC}. Looking at the $\lambda_\mathsf{max}$ in the `Initial' row for all 6 attacks (details are in the supplementary material), it can be observed that all of these backdoor models optimizes to a sharp minima. As these models are optimized on both distributions, they also have high attack success rates (ASR) as well as high clean test accuracy (ACC). Note that, the measure of smoothness is done \emph{w.r.t.} clean data distribution. 
The use of clean distribution in our smoothness analysis is driven from the practical consideration as our particular interest lies with the performance \emph{w.r.t.} clean distribution; more details are in \emph{the supplementary material}.
Since high ASR and ACC indicate that the model had learned both distributions, it supports Conjecture~1. 

\medskip

\textbf{Conjecture 2.} Through \emph{proper} fine-tuning with clean validation data, a backdoor model can be re-optimized to a smoother minima \emph{w.r.t.} clean data distribution. Optimization to a smoother minima leads to backdoor purification, \ie low ASR and high ACC.

\medskip
By \emph{proper fine-tuning}, we imply that the fine-tuning will lead to an optimal solution \emph{w.r.t.} the data distribution we fine-tune the model with. 
To support Conjecture 2, we show the removal performances of fine-tuning based purification methods in Table~\ref{tab:eigen-ASR-ACC}. To remove backdoor using a clean validation set ($\sim$1\% of  train-set), we fine-tune different parts of the DNN for 100 epochs with a learning rate of 0.01. As shown in Table~\ref{tab:eigen-ASR-ACC}, after proper fine-tuning (Full-Net, CNN-Bbone), the backdoor model re-optimizes to a smoother minima that leads to successful backdoor removal.   

\medskip

\noindent\textbf{One-Layer Fine-tuning.} We observe that one can remove the backdoor by fine-tuning either the full network or only the CNN backbone (using SGD). However, these methods can be computationally costly and less practical. Furthermore, such fine-tuning often leads to high drop in ACC. As an alternative, one could fine-tune only the last or classification (Cls.) layer. However, even with a small validation set, a one-layer network becomes a shallow network to optimize. According to the spin-glass analogy in \cite{choromanska2015loss}, as the network size decreases the probability for the SGD optimizer to find \emph{sharp local minima or poor quality minima} increases accordingly. In case of shallow network, the quality of minima is decided by their distances from the global minima. \cite{choromanska2015loss} also observes that the process of finding a path from bad local minima to a good quality solution or global minima takes \emph{exponentially long time}. Therefore, it is not always feasible to use the SGD optimizer for shallow network. Table~\ref{tab:eigen-ASR-ACC} (row--Cls. (SGD)) corroborates this hypothesis as SGD optimizer fails to escape the sharp minima resulting in similar ASRs as the initial backdoor model. Instead of using SGD, one can use natural gradient descent (NGD) that has \emph{higher probability of escaping the bad local minima as well as faster convergence rate}, specifically in the shallow network scenario \cite{amari1998natural,martens2015optimizing}. Therefore, to effectively purify a backdoor model, we propose a novel Fisher Information matrix based backdoor purification objective function and optimize it using the NGD optimizer.

\section{Natural Gradient Fine-tuning (NGF)} \label{method}
This section presents our proposed backdoor purification method---Natural Gradient Fine-tuning (NGF). Recall that the backdoor model under consideration is $f_\theta (.)$, where $\theta$ is the model parameter.
Let us decompose $\theta$ as,
\[
    \theta =  \{\W_{0,1},\W_{1,2}, \W_{2,3}, \cdots, \W_{L-1,L}\},
\]  
where $\W_{i,i+1}$ is the parameters between layer $i$ and layer $i+1$, commonly termed as $(i+1)^{th}$ layer's parameters. $\W_{L-1, L}$ is the $L^{th}$ layer's (Cls. layer) parameters, and we are particularly interested in fine-tuning only this layer.
Now, consider a validation set, $\mathbb{D}_{\mathsf{val}} =  \{X_{\mathsf{val}}, Y_{\mathsf{val}}\}$ that contains only clean samples. We denote $\theta_L = \W_{L-1,L}$ 
as the $L^{th}$ layer's parameters\footnote{Notice that $\theta_L$ is a \textbf{vector} flattening the $L^{th}$ layer's parameter.} and $\theta_{L,i}$ is the $i^{th}$ element of $\theta_L$. To purify the backdoor model, we formulate the following loss  
\begin{equation}\label{eqn:loss2}
    \mathcal{L}_p(y,f_\theta(x)) = \mathcal{L}(y,f_\theta(x)) + \frac{\eta}{2}\sum_{\forall i} \mathsf{diag} (F(\bar{\theta}_L))_i \cdot (\theta_{L,i} - \bar{\theta}_{L,i})^2,
\end{equation}
which is a combination of the CE loss on the validation set and a regularizer. Here, $\bar{\theta}_L$ is $L^{th}$ layer parameters of the initial backdoor model, \ie $\theta^{(0)}_L = \bar{\theta}_L$ and remains fixed throughout the purification phase.

In a backdoor model, some neurons/parameters are more vulnerable than others. The vulnerable parameters are believed to be the ones that are sensitive to poison/trigger data distribution \citep{wu2021adversarial}. In general, CE loss does not discriminate whether a parameter is more sensitive to clean or poison distribution. Such lack of discrimination may allow drastic/unwanted changes to the parameters responsible for learned clean distribution. This usually leads to sub-par clean test accuracy after purification and it requires additional measures to fix this issue. Motivated by \cite{kirkpatrick2017overcoming}, we introduce a \textit{clean distribution aware regularization} term as a product of two terms: i) an error term that accounts for the deviation of $\theta_L$ from $\bar{\theta}_L$; 
ii) a vector, $\mathsf{diag}(F(\bar{\theta}_L))$, consisting of the diagonal elements of FIM ($F(\bar{\theta}_L)$). As the first term controls the changes of parameters \emph{w.r.t.} $\bar{\theta}_L$, it helps the model to remember the already learned distribution. However, learned data distribution consists of both clean and poison distribution. To explicitly force the model to remember the \emph{clean distribution}, we compute $F(\bar{\theta}_L)$ using a \emph{clean} validation set; with similar distribution as the learned clean data. Note that, $\mathsf{diag}(F(\bar{\theta}_L))_i$ represents the square of the derivative of log-likelihood of clean distribution \emph{w.r.t.} $\bar{\theta}_{L,i}$, $[\nabla_{\bar{\theta}_{L,i}}\mathsf{log}~p(y|x,\theta)]^2$~(ref. eqn.~(\ref{F})). In other words, $\mathsf{diag}(F(\bar{\theta}_L))_i$ is the measure of importance of $\bar{\theta}_{L,i}$ towards remembering the learned clean distribution. If $\mathsf{diag}(F(\bar{\theta}_L))_i$ has a higher importance, we allow minimal changes to $\bar{\theta}_{L,i}$ over the purification process.
This careful design of such a regularizer improves the clean test performance significantly. We use $\eta$ as a regularization constant.

\begin{algorithm}[t]
	\DontPrintSemicolon
	\scriptsize
    \caption{Natural Gradient Fine-tuning (NGF)}
    \textbf{Input:} Backdoor Model ($f_\theta$(.)), $1\%$ Clean Validation Set $\mathbb{D}_{val}$, Number of Purification Epochs $\mathcal{N}$ \\ 
    Initialize all mask values in $M_0$ as 1 \\
    $\mathcal{X}, \mathcal{Y} \leftarrow \mathbb{D}_{val}$ 
    
    $\displaystyle F(\bar{\theta}_L) \leftarrow \frac{1}{|\mathbb{D}_{val}|}\sum_{x\in\mathcal{X},y\in\mathcal{Y}}\left[\nabla_{\bar{\theta}_L}~\mathsf{log} ~p(y|x,\theta)\cdot\left(\nabla_{\bar{\theta}_L}~\mathsf{log} ~p(y|x,\theta)\right)^T\right]$\tcp*{$\bar{\theta}_L$ is the last layer's parameter of the initial backdoor model.}
    
    \For{$i=1$ \KwTo $\mathcal{N}$}
        {   
            $\displaystyle \mathcal{L} = \mathcal{L}_{CE}(\mathcal{Y}, f_{\theta^{(i)}}(\mathcal{X})) + \frac{\eta}{2}\sum_j(\mathsf{diag}(F(\bar{\theta}_L)))_j\cdot (\theta_{L,j}^{(i)} - \bar{\theta}_{L,j})^2$\tcp*{the superscript $i$ in $\theta^{(i)}$ denotes the parameter of $i^{th}$ iteration}
            
            $\displaystyle F \leftarrow \frac{1}{|\mathbb{D}_{val}|}\sum_{x\in\mathcal{X},y\in\mathcal{Y}}\left[\nabla_{\theta_L^{(i)}}~\mathsf{log} ~p(y|x,\theta^{(i)})\cdot\left(\nabla_{\theta_L^{(i)}}~\mathsf{log} ~p(y|x,\theta^{(i)})\right)^T\right]$\tcp*{$\theta_L^{(i)}$ is the last layer's parameter at $i^{th}$ iterations}
            
            $\theta_L^{(i+1)} \leftarrow \theta_L^{(i)} - \alpha \cdot F^{-1} \nabla_{\theta_L^{(i)}} (\mathcal{L}) $ \tcp*{$\alpha$ is the learning rate}
            
            $\theta^{(i+1)} \leftarrow \{\W_{0,1},\W_{1,2},  \cdots , \W_{L-2,L-1}, \theta_L^{(i+1)}\}$ \tcp*{$\W_{i,i+1}$'s are frozen parameters}
        }
        
        $\theta_p \leftarrow \{\W_{0,1},\W_{1,2},  \cdots , \W_{L-2,L-1}, \theta_L^{(\mathcal{N})}\} $ \tcp*{$\theta_p$ is the purified model's parameter}
        
    \textbf{Output:} Purified Model, $f_{\theta_p}$
\label{alg:main_algorithm}
\end{algorithm}

\begin{table*}[t]
\footnotesize
\centering
 \caption{Comparison of different defense methods for four benchmark datasets. Backdoor removal performance, \ie~drop in ASR, against a wide range of attacking strategies show the effectiveness of NGF. For CIFAR10 and GTSRB, the poison rate is $10\%$. For Tiny-ImageNet and ImageNet, we employ ResNet34 and ResNet50 architectures, respectively. We use a poison rate of 5\% for these 2 datasets and report performance on successful attacks (ASR close to 100\%) only.  
Average drop ($\downarrow$) indicates the \% changes in ASR/ACC compared to the baseline, \ie ASR/ACC of \emph{No Defense}. Higher ASR drop and lower ACC drop is desired for a good defense.}
\scalebox{0.95}{
\begin{tabular}{c|c|cc|cc|cc|cc|cc|cc}
\toprule
\multirow{2}{*}{Dataset} & Method & \multicolumn{2}{c|}{\begin{tabular}[c|]{@{}c@{}}No Defense\end{tabular}} & \multicolumn{2}{c|}{Vanilla FT} & \multicolumn{2}{c|}{ANP} & \multicolumn{2}{c|}{I-BAU} & \multicolumn{2}{c|}{AWM} &  \multicolumn{2}{|c}{NGF (Ours)}\\ \cmidrule{2-14}
 & Attacks & ASR &ACC & ASR &ACC & ASR &ACC & ASR &ACC & ASR &ACC & ASR &ACC \\ \cmidrule{1-14}
\multirow{12}{*}{\footnotesize CIFAR-10} 
  &  \emph{Benign} & 0 & 95.21 & 0 & 92.28 & 0 & 93.98&0&93.56&0&93.80 & 0 & \textbf{94.10}  \\
 &   Badnets & 100 & 92.96 & 4.87 & 85.92 & 2.84 & 85.96&9.72&87.85&4.34&86.17 & \textbf{1.86} &\textbf{ 88.32} \\
 &   Blend & 100 & 94.11  & 4.77 & 87.61 & 3.81 & 89.10&11.53&90.84&2.13&88.93 &\textbf{ 0.38} & \textbf{91.17} \\
  &  Troj-one & 100 & 89.57& 3.78 & 82.18& 5.47 & {85.20} & 7.91&\textbf{87.24}&5.41&86.45  &\textbf{2.64} & 84.21 \\
  &  Troj-all & 100 & 88.33 & 3.91 & 81.95& 5.53 & 84.89 & 9.82&85.94&4.42&84.60 & \textbf{2.79} & \textbf{86.10} \\
 &   SIG & 100 & 88.64 & 1.04 & 81.92 & 0.37 & 83.60 &4.12&83.57&0.90&83.38& \textbf{ 0.12} & \textbf{84.16}  \\
  &  Dyn-one & 100 & 92.52 & 4.73 & 88.61 & 1.78 & 86.26 &10.48&89.16&3.35&88.41& \textbf{1.17} & \textbf{90.97} \\
 &  Dyn-all & 100 & 92.61 & 4.28 & 88.32 & 2.19 & 84.51 &10.30&89.74&2.46&87.72& \textbf{1.61} & \textbf{90.19}\\
 &  CLB & 100 & 92.78 & 1.83 & 87.41 & 1.41 & 85.07 &5.78&86.70&1.89&84.18& \textbf{1.04} & \textbf{88.37}\\
 &  CBA & 93.20 & 90.17 & 27.80 & 83.79 &45.11&85.63&36.12&85.05&38.81&85.58&\textbf{24.60}&\textbf{85.97}\\
 & FBA & 100&90.78 &7.95 &82.90 &66.70&\textbf{87.42}&10.66&87.35&22.31&87.06&\textbf{6.21}&86.96\\
 & WaNet&98.64&92.29&5.81&86.70&3.18&89.24&10.72&85.94&2.96&89.45&\textbf{2.38}&\textbf{89.65}\\
 & ISSBA&99.80&92.80&6.76&85.42&\textbf{3.82}&89.20&12.48&90.03&4.57&89.59&4.24&\textbf{90.18}\\
& BPPA & 99.70 & 93.82 & 9.94 & 90.23 & 10.46&90.57&9.94&90.68&10.60&90.88&\textbf{7.14}&\textbf{91.84}\\
 \cmidrule{2-14} 
 & {Avg. Drop} & - &-  &  $92.61\downarrow$ & $6.03\downarrow$ &  $87.59\downarrow$ &  $4.98\downarrow$ &$87.82\downarrow$&$3.95\downarrow$&$91.32\downarrow$&$4.53\downarrow$& {\textbf{95.01}} $\downarrow$ &{\textbf{3.33}} $\downarrow$ \\ 
 \cmidrule{1-14}
\multirow{9}{*}{\footnotesize GTSRB} 
  & \emph{Benign} & 0 & 97.87 & 0&93.08 & 0 & 95.42&0&96.18&0&95.32 & 0 & \textbf{95.76}\\
  & Badnets & 100 & 97.38 & 1.36&88.16 & 0.35 & 93.17 &2.72&\textbf{94.55}&2.84&93.58& \textbf{0.24}  & {94.11} \\
  & Blend & 100 & 95.92 & 5.08&89.32 & 4.41 & 93.02 &4.13&\textbf{94.30}&4.96&92.75& \textbf{2.91} & {93.31}  \\
  & Troj-one & 99.50 & 96.27 & 2.07&90.45 & 1.81 & 92.74&3.04&93.17&2.27&93.56 & \textbf{ 1.21} &  \textbf{94.18} \\
  &  Troj-all & 99.71 & 96.08 & 2.48&89.73& 2.16 & 92.51 &2.79&93.28&1.94&92.84& \textbf{1.58} & \textbf{93.87} \\
  &  SIG & 97.13 & 96.93 & \textbf{1.93} &91.41 & 6.17 & 91.82 &2.64&93.10&5.32&92.68 & 3.24 & \textbf{93.48}  \\
  &  Dyn-one & 100 & 97.27 & 2.27&91.26 & 2.08 & 93.15 &5.82&\textbf{95.54}&1.89&93.52&\textbf{1.51} & {94.27}  \\
  &  Dyn-all & 100 & 97.05 & 2.84 & 91.42 & 2.49 & 92.89 &4.87&93.98&2.74&93.17& \textbf{1.26} & \textbf{94.14}\\
  & BPPA & 99.18 & 98.12 & 5.14 & 94.48 & 7.19 & 93.79&8.63&94.50&5.43&94.22 &\textbf{4.45}&\textbf{95.27} \\\cmidrule{2-14}  
& {Avg. Drop} & - & -  & 96.54 $\downarrow$ & 6.10 $\downarrow$ & 96.10$\downarrow$ & 3.99 $\downarrow$  &$95.11\downarrow$&$2.83\downarrow$&$96.02\downarrow$&$3.59\downarrow$& {\textbf{97.39}} $\downarrow$ &{\textbf{2.79}} $\downarrow$\\ 
\cmidrule{1-14}
\multirow{7}{*}{\footnotesize Tiny-ImageNet} 
&\emph{Benign}&0&62.56&0&58.20&0&59.29&0&59.34&0&59.08&0&\textbf{59.67}\\
&Badnets&100&59.80&3.84&53.58&61.23&55.41&13.29&54.56&31.44&54.81&\textbf{2.34}&\textbf{55.84}\\
&Trojan&100&59.16&6.77&52.62&79.56&54.76&11.94&\textbf{55.10}&38.23&54.28&\textbf{3.38}&54.87\\
&Blend&100&60.11&2.18&51.22&81.58&54.70&17.42&54.19&41.37&53.78&\textbf{1.58}&\textbf{54.98}\\
&SIG&98.48&60.01&5.02&52.18&28.67&54.71&9.31&\textbf{55.72}&27.68&54.11&\textbf{2.81}&54.63\\
&CLB&97.71&60.33&5.61&51.68&16.24&55.18&10.68&54.93&36.52&55.02&\textbf{4.06}&\textbf{55.40} \\\cmidrule{2-14}  
& {Avg. Drop} & - & -  & 94.55 $\downarrow$ & 7.63 $\downarrow$ & 45.38$\downarrow$ & 4.93 $\downarrow$ &$86.71\downarrow$&$4.98\downarrow$&$64.19\downarrow$&$5.48\downarrow$ & {\textbf{96.40}} $\downarrow$ &{\textbf{4.74}} $\downarrow$\\ 
\cmidrule{1-14}
\multirow{7}{*}{\footnotesize ImageNet} 
&\emph{Benign}&0&77.06&0&73.52&0&68.85&0&74.21&0&71.63&0&\textbf{74.51}\\
&Badnets&99.24&74.53&5.91&69.37&43.31&66.28&21.87&69.46&21.18&69.44&\textbf{4.61}&\textbf{70.46}\\
&Trojan&99.21&74.02&4.63&69.15&38.81&66.14&25.74&69.35&28.85&68.62&\textbf{4.02}&\textbf{69.97}\\
&Blend&100&74.42&4.43&70.20&57.79&65.51&27.45&68.61&34.15&68.91&\textbf{3.83}&\textbf{70.52}\\
&SIG&94.66&74.69&3.23&69.82&16.28&66.08&15.37&70.02&16.47&69.74&\textbf{2.94}&\textbf{71.36}\\
&CLB&95.06&74.14&3.71&69.19&18.37&66.41&21.64&69.70&23.50&69.32&\textbf{3.05}&\textbf{70.25}\\
&Dynamic&95.06&74.14&3.71&69.19&18.37&66.41&21.64&69.70&23.50&69.32&\textbf{3.05}&\textbf{70.25}\\
\cmidrule{2-14}  
& {Avg. Drop} & - & -  & 93.25 $\downarrow$ &  4.81$\downarrow$ & 62.72$\downarrow$ & 8.28 $\downarrow$ &$75.22\downarrow$&$4.93\downarrow$&$72.80\downarrow$& $5.15\downarrow$& {\textbf{93.94}} $\downarrow$ &{\textbf{3.85}} $\downarrow$\\ 
\bottomrule
\end{tabular}}
\vspace{-3mm}
 \label{tab:main}
\end{table*}

The overall optimization problem using the loss-function defined in (\ref{eqn:loss2}) for purifying the backdoor model $f_\theta$ is as follows: 

\medskip
\noindent Objective function: 
\begin{equation}
  \theta_p := \argmin_{\theta_{L}} \mathcal{L}_p (y, f_\theta(x)); ~~ x\in X_{val},~ y\in Y_{val}  
\end{equation}
Update Policy:
\begin{equation}
    \theta_L^{(t+1)} \leftarrow \theta_L^{(t)} - \alpha F(\theta_L^{(t)})^{-1}\nabla_{\theta_L} \mathcal{L}_p \label{eq:F-1}\\
\end{equation}
Where,
\begin{equation}
    F(\theta_L)= \frac{1}{n}\sum_{j=1}^n \left(\nabla_{\theta_L}\mathsf{log}~p(y_j|x_j,\theta)\cdot(\nabla_{\theta_L}\mathsf{log}~p(y_j|x_j,\theta))^T\right)\label{F}
\end{equation}

Here, $F\in \mathbb{R}^{|\theta_L|\times|\theta_L|}$ is the FIM, and $n$ is the validation set size. Notice that, as we only consider fine-tuning of $L^{th}$-layer, the computation of $F$ and $F^{-1}$ ($|\theta_L|\times|\theta_L|$ matrices) becomes tractable. 
After solving the above optimization problem, we will get modified parameters, $\overline{\W}_{L-1,L}$. Finally, we get the purified model, $f_{\theta_p}$ with $\theta_p$ as
\begin{align*}
    \theta_p =  \{\W_{0,1}, \W_{1,2} , \W_{2,3} , \cdots, \overline{\W}_{L-1,L}\}
\end{align*}

Fig.~\ref{fig:ls-pure-sgd} and~\ref{fig:ls-pure-ngd} show that NGF indeed does reach the smooth minima as opposed to SGD based fine-tuning.
We provide additional results in Table~\ref{tab:eigen-ASR-ACC} for both NGF and SGD. Notice that the purified model seems to have a smoother loss surface than the benign model (2.7 vs. 20.1 for $\lambda_\mathsf{max}$). This, however, does not translate to better ACC than the benign model. The ACC of the purified model is always bounded by the ACC of the backdoor model. To the best of our knowledge, our study on the correlation between loss-surface smoothness and backdoor purification is novel. NGF is also the first method to employ a second-order optimizer for purifying backdoor. \textit{More details are in the supplementary material.}
The manner in which we perform natural gradient fine-tuning is described in Algorithm~\ref{alg:main_algorithm}. 
After purification, the model should behave like a benign/clean model producing the same prediction irrespective of the presence of the trigger.
\section{Experimental Results}\label{sec:experiment}
\subsection{Evaluation Settings}
\textbf{Datasets:} To begin with, we evaluate our proposed method through conducting a wide range of experiments on two widely used datasets for backdoor attack study: {\bf{CIFAR10}}~\citep{krizhevsky2009learning} with 10 classes, {\bf{GTSRB}}~\citep{stallkamp2011german} with 43 classes. As a test of scalability, we also consider {\bf{Tiny-ImageNet}}~\citep{le2015tiny} with 100,000 images distributed among 200 classes and {\bf{ImageNet}}~\citep{deng2009imagenet} with 1.28M images distributed among 1000 classes. 

\begin{figure*}[t]
\centering
\begin{subfigure}{0.245\linewidth}
    \includegraphics[width=1\linewidth]{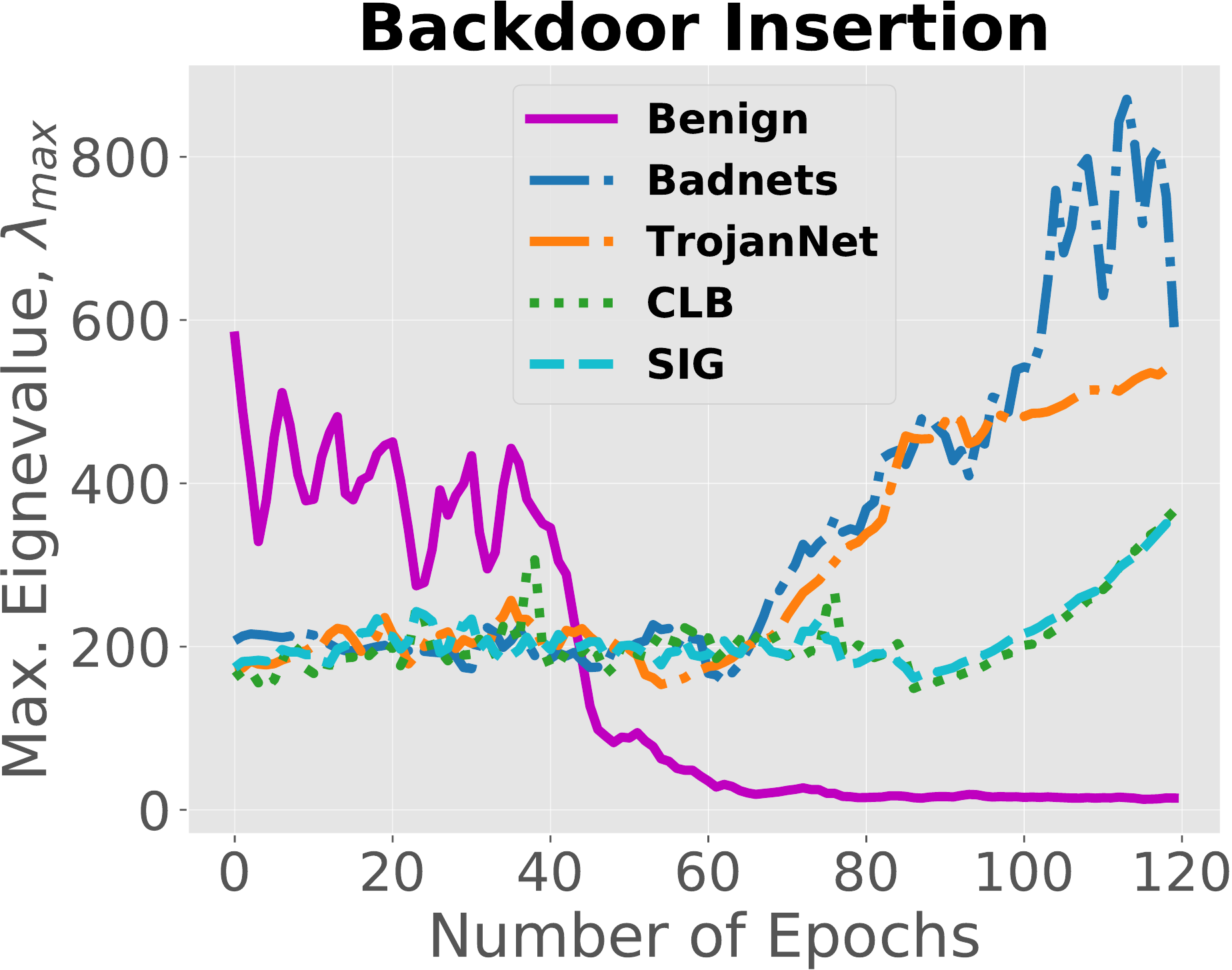}
    \caption{\scriptsize $\lambda_\mathsf{max}$ vs. Epochs}
    \label{fig:backdoor_eigens}
\end{subfigure}
\hfill
\begin{subfigure}{0.245\linewidth}
    \includegraphics[width=1\linewidth]{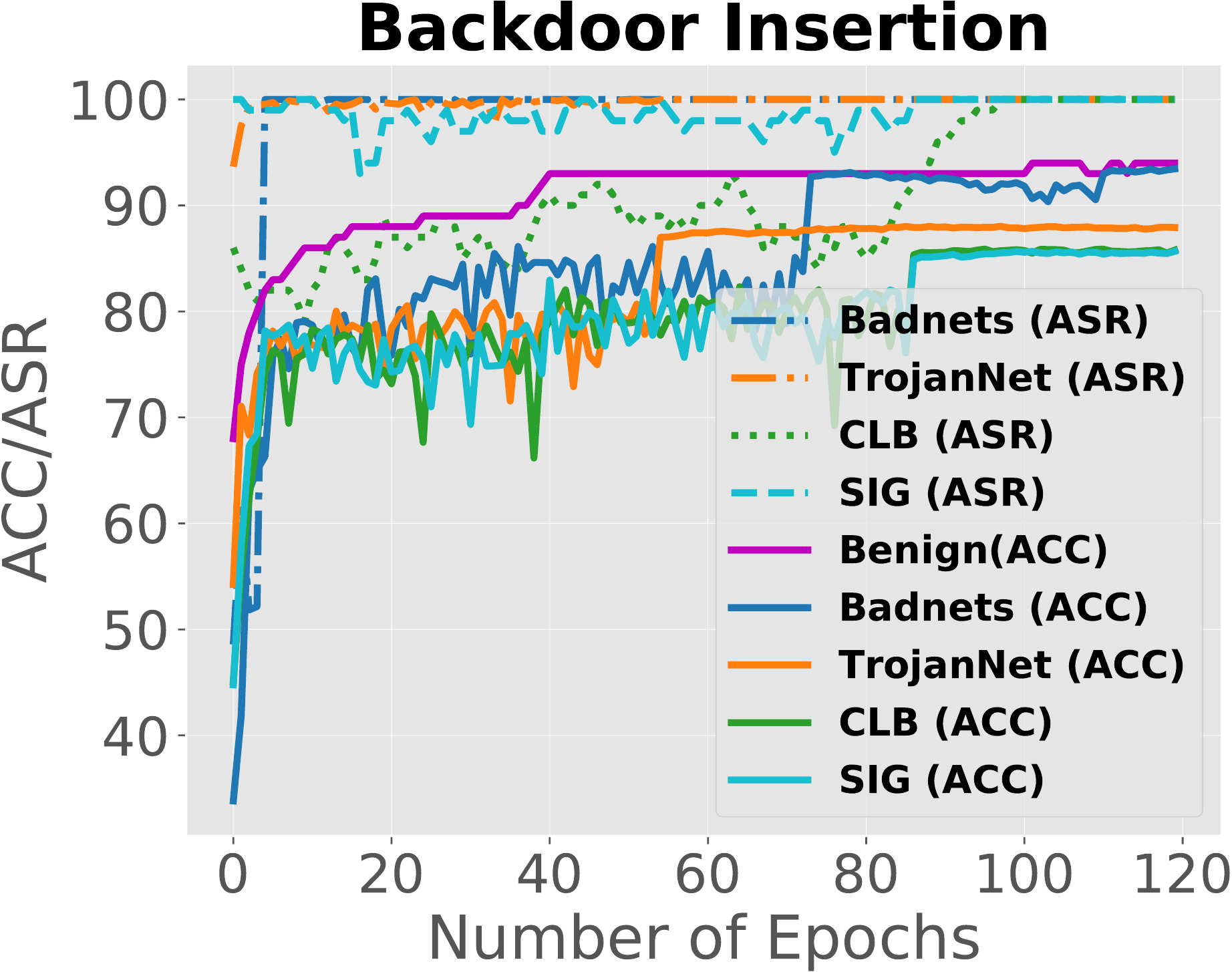}
    \caption{\scriptsize  ACC/ASR vs. Epochs}
    \label{fig:backdoor_asr}
\end{subfigure}
\hfill
\begin{subfigure}{0.245\linewidth}
    \includegraphics[width=1\linewidth]{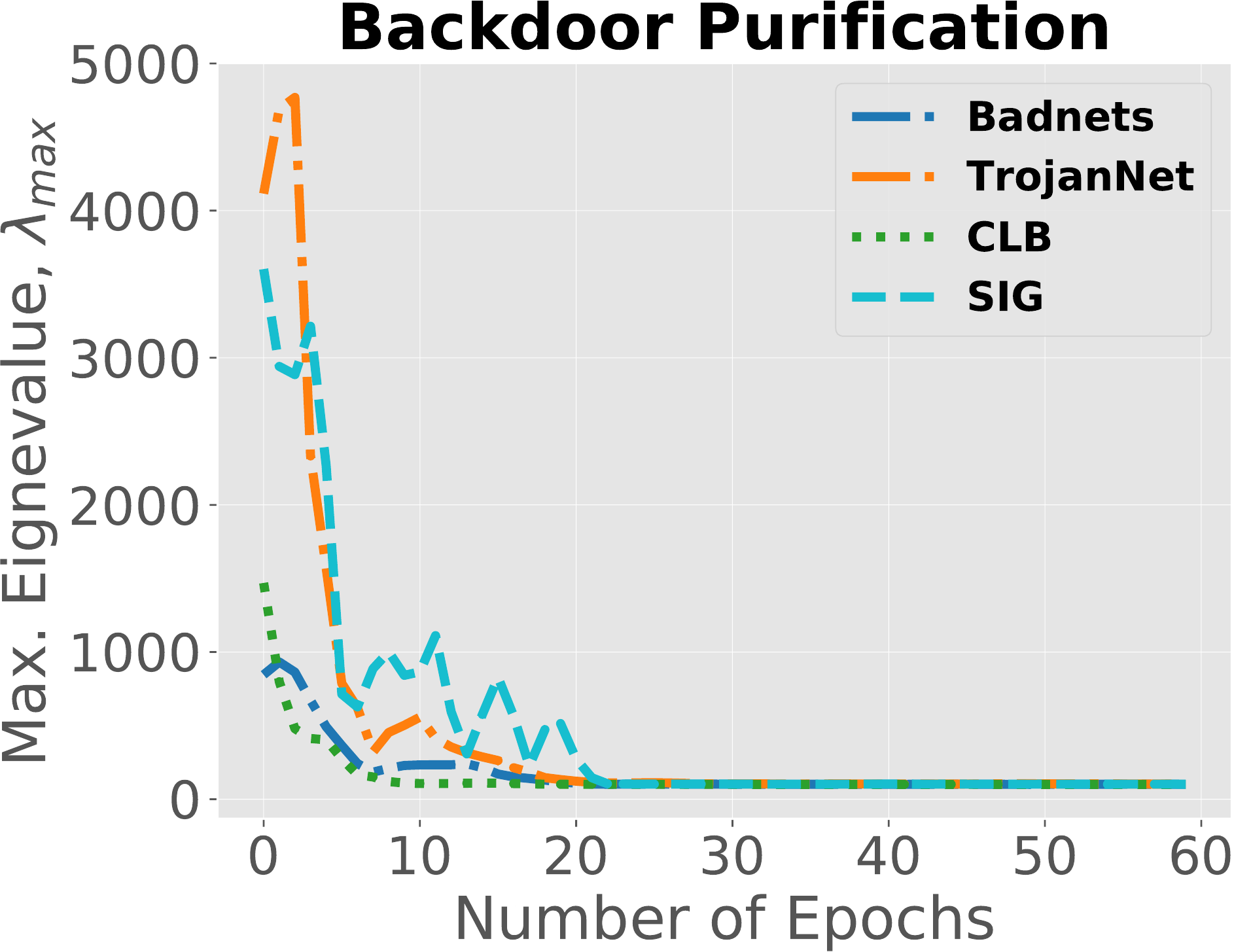}
    \caption{\scriptsize $\lambda_\mathsf{max}$ vs. Epochs}
    \label{fig:purified_eigens}
\end{subfigure}
\hfill
\begin{subfigure}{0.24\linewidth}
    \includegraphics[width=1\linewidth]{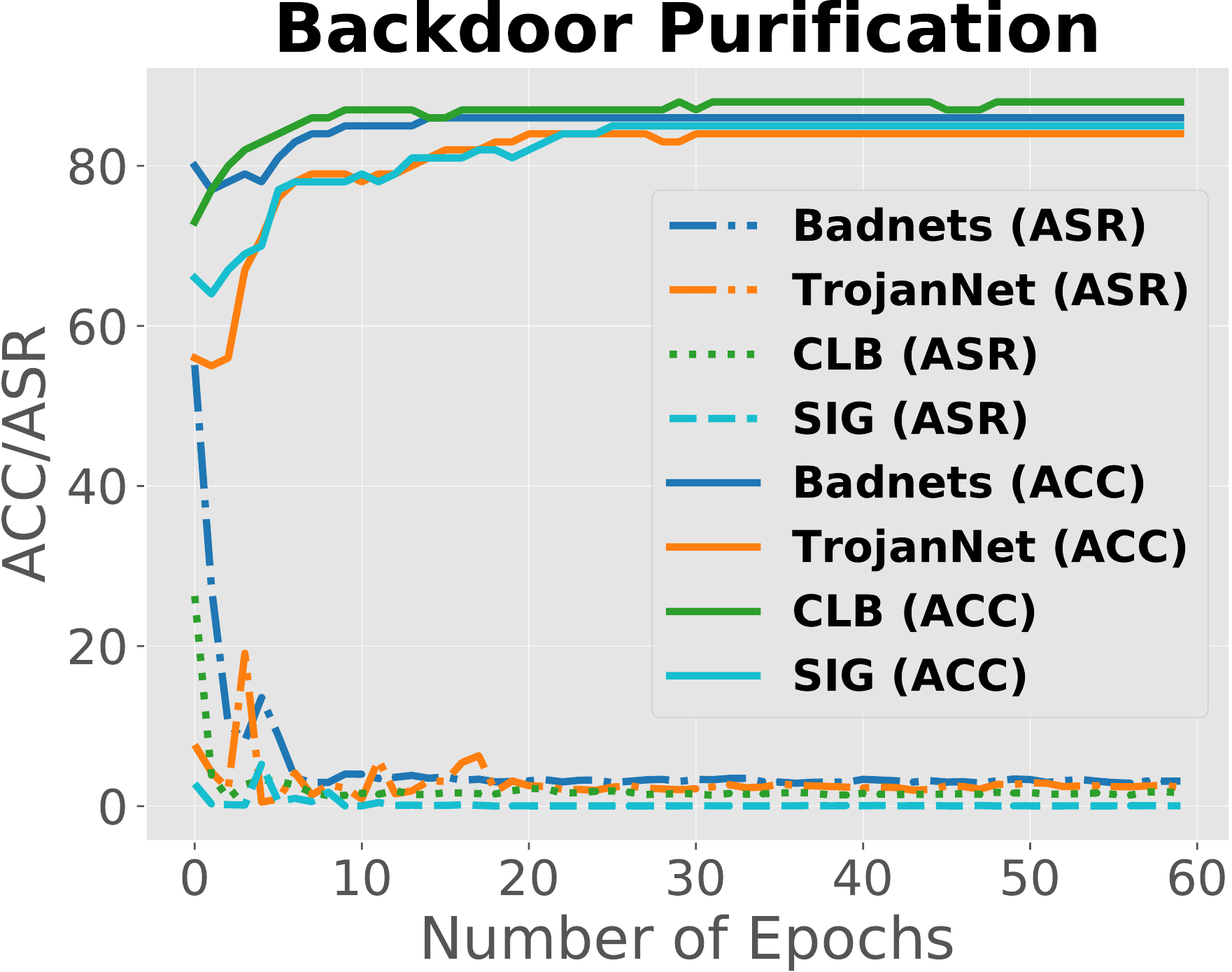}
    \caption{\scriptsize ACC/ASR vs. Epochs}
    \label{fig:purified_asr}
\end{subfigure}
\caption{Loss Surface characteristics of a DNN during backdoor insertion and purification processes. a \& b) As the joint optimization on clean and poison distribution progresses, \ie high ACC \& ASR, the loss surface becomes less and less smoother, \ie high $\lambda_\mathsf{max}$). c \& d) One can purify  backdoor by gradually making the loss surface smoother. We use CIFAR10 dataset with four different attacks.}
\end{figure*}
\begin{table*}
    \centering
    \caption{ Performance comparison of NGF to other SGD-based optimizers. A more suitable sharpness-aware SGD-based optimizer is also considered here. However, NGF is far more effective in purifying backdoor (lower ASR) due to its consistent convergence to smooth minima. We use CIFAR10 dataset for these evaluations.}
    \scalebox{1}{
    \begin{tabular}{c|cc|cc|cc|cc|cc|cc}
    \toprule
    Defense & \multicolumn{2}{c|}{No Defense }   & \multicolumn{2}{c|}{AdaGrad} & \multicolumn{2}{c|}{RMSProp}& \multicolumn{2}{c|}{Adam} & \multicolumn{2}{c|}{SAM} & \multicolumn{2}{c}{NGF (Ours)} \\
    \midrule
    Attacks & ASR & ACC & ASR & ACC & ASR & ACC & ASR & ACC & ASR & ACC & ASR & ACC \\
    \midrule
    Badnets & 100 & 92.96 & 96.54 & 91.16 & 98.33 & \textbf{91.73} & 97.68 &91.45  & 91.08& 90.12 & \textbf{1.86} & 88.32\\
    Blend & 100 & {94.11} & 97.43 & 91.67 & 95.41 & \textbf{92.21} & 94.79 & 92.15 & 89.25 & 91.11 & \textbf{0.38} & 91.17\\
    Trojan & 100 & 89.57 & 95.52&\textbf{88.51} & 94.87 & 88.02 & 96.74 & 87.98 & 92.15 & 88.33 & \textbf{2.64} & 84.21\\
    Dynamic & 100 & 92.52 & 97.37 & \textbf{91.45} & 93.50 & 91.12 & 96.90 & {91.40} & 92.24 & 90.79 & \textbf{1.17} & 90.97 \\
    SIG & 100 & 88.64   &86.20 &87.98 &86.31 &87.74&85.66 &87.75&81.68 &\textbf{88.04} & \textbf{0.31} & 83.14  \\
    CLB & 100 & 92.78  & 96.81 & 90.86&95.53 & 90.96 & 95.87 &\textbf{91.02} & 91.04 & 90.97 & \textbf{1.04} & 88.37 \\
    \bottomrule
    \end{tabular}
    }
    \label{tab:different_optimizer}
\end{table*}
\noindent\textbf{Attacks Configurations:} We consider 13 state-of-the-art backdoor attacks: 1) Badnets~\citep{gu2019badnets}, 2) Blend attack~\citep{chen2017targeted}, 3 \& 4) TrojanNet (Troj-one \& Troj-all)~\citep{liu2017trojaning}, 5) Sinusoidal signal attack (SIG)~\citep{barni2019new}, 6 \& 7) Input-Aware Attack (Dyn-one and Dyn-all)~\citep{nguyen2020input}, 8) Clean-label attack (CLB) ~\citep{turner2018clean}, 9) Composite backdoor (CBA)~\citep{lin2020composite}, 10) Deep feature space attack (FBA)~\citep{cheng2021deep}, 11) Warping-based backdoor attack (WaNet)~\citep{nguyen2021wanet}, 12) Invisible triggers based backdoor attack (ISSBA)~\citep{li2021invisible}, and 13) 
Quantization and contrastive learning based attack (BPPA)~\citep{wang2022bppattack}. To ensure fair comparison, we follow the similar trigger patterns and settings as in their original papers. In Troj-one and Dyn-one attacks, all of the triggered images have same target label. On the other hand, target labels are uniformly distributed over all classes for Troj-all and Dyn-all attacks. For creating these attacks on CIFAR10 and GTSRB, we use a poison rate of 10\% and train a PreActResNet18~\citep{he2016identity} and a WideResNet-16-1~\citep{zagoruyko2016wide} architectures, respectively, for 250 epochs with an initial learning rate of 0.01. 
More details on hyper-parameters and overall training settings can be found in \emph{the supplementary material}. 

\begin{table}
\vspace{1mm}
    \caption{Avg. runtime comparison for different datasets. Here, \#Parameters is the total number of parameters in the last layer. An NVIDIA RTX 3090 GPU is used for all experiments. }
    \centering
    \scalebox{1}{
    \begin{tabular}{c|c|c|c}
    \toprule
    Dataset & \# Parameters & Method  & Runtime (Sec.) \\ 
    \midrule
     \multirow{2}{*}{CIFAR10} & \multirow{2}{*}{5120} & FT & 78.1 \\ 
     & & NGF & \textbf{38.3} \\  \cmidrule{1-4}
    \multirow{2}{*}{GTSRB} & \multirow{2}{*}{22016} & FT & 96.2 \\
     & & NGF & \textbf{47.4} \\ \cmidrule{1-4}
    \multirow{2}{*}{Tiny-ImageNet} & \multirow{2}{*}{409.6K} & FT & 637.6  \\
     & & NGF & \textbf{374.2} \\ \cmidrule{1-4}
     \multirow{2}{*}{ImageNet} & \multirow{2}{*}{2.048M}& FT & 2771.6 \\
     & & NGF & \textbf{1681.4} \\ 
     \bottomrule
    \end{tabular}}
    \vspace{-1mm}
    \label{tab:runtime_dataset}
\end{table}
\noindent\textbf{Defenses Configurations:} We compare our approach with 4 existing backdoor mitigation methods:  1) Vanilla Fine-Tuning (FT); where we fine-tune all DNN parameters, 2) Adversarial Neural Pruning (ANP)~\citep{wu2021adversarial} with $1\%$ clean validation data, 3) Implicit Backdoor Adversarial Unlearning (I-BAU)~\citep{zeng2021adversarial}, 4) Adversarial Weight Masking (AWM)~\citep{chai2022one}, 5) Fine-Pruning (FP) ~\cite{liu2017neural}, 6) Mode Connectivity Repair (MCR)~\cite{zhao2020bridging}, and 7) Neural Attention Distillation (NAD)~\cite{li2021neural}. However, we move the experimental results for defenses 5, 6, and 7 to the supplementary material due to the page limitation.
To apply NGF on CIFAR10, we fine-tune the last layer of the DNN for $E_p$ epochs with $1\%$ clean validation data. Here, $E_p$ is the number of purification epochs and we choose a value of 100 for this. For optimization, we choose a learning rate of 0.01 with a decay rate of 0.1/40 epochs and consider regularization constant $\eta$ to be 0.1. Additional experimental details for NGF and other defense methods are in \emph{the supplementary material}. For GTSRB, we increase the validation size to $3\%$ as there are less samples available per class. Rest of the training settings are same as CIFAR10. For NGF on Tiny-ImageNet, we consider a validation size of 5\% as a size less than this seems to hurt clean test performance (after purification). We fine-tune the model for 15 epochs with an initial learning rate of 0.01 with a decay rate of 0.3/epoch. 
Finally, we validate the effectiveness of NGF on ImageNet. For removing the backdoor, we use 3\% validation data and fine-tune for 2 epochs. A learning rate of 0.001 has been employed with a decay rate of 0.005 per epoch. 
\emph{We define the effectiveness of a defense method in terms of average drop in ASR and ACC over all attacks. A highly effective method should have a high drop in ASR with a low drop in ACC.}
We define ASR as the percentage of poison test samples that are classified to the adversary-set target label. 
 \begin{table*}[t]
    \centering
    \caption{ Performance of NGF while fine-tuning all layers of DNN. We also consider SAM and SGD based fine-tuning of all layers here. The results shown here are for CIFAR10 dataset.} 
    \vspace{-2mm}
    \scalebox{0.9}{
    \begin{tabular}{c|cc|cc|cc|cc|cc|cc|cc|c}
    \toprule
    \multirow{2}{*}{Methods} & \multicolumn{2}{c|}{Badnets} & \multicolumn{2}{c|}{Blend} & \multicolumn{2}{c|}{Trojan} & \multicolumn{2}{c|}{Dynamic} & \multicolumn{2}{c|}{CLB}& \multicolumn{2}{c|}{SIG} & \multicolumn{2}{c|}{CBA} & Runtime \\ 
      &  ASR & ACC & ASR & ACC& ASR & ACC& ASR & ACC&  ASR & ACC& ASR & ACC & ASR & ACC & (Secs.) \\
    \midrule
    Initial&100&92.96&100&94.11&100&89.57&100&92.52&100&92.78&100&88.64 & 93.20 & 90.17& -- \\
    SGD (All Layers) & 4.87 & 85.92 & 4.77 & 87.61 & 3.78 & 82.18 & 4.73 & 88.61 & 1.83 & 87.41 & 1.04 & 81.92 & 27.80 & 83.79 & 78.1  \\
    SAM (All Layers) & 3.91 & 85.75	& 2.74 & 88.26  & 	3.53 & 82.52 & 3.28 & 87.04 & 1.47 & 86.30  & 0.38 & 84.70 & 26.14 & 85.41 & 116.3 \\
    \rowcolor{aliceblue} NGF (Last layer) & 1.86 & 88.32&\textbf{0.38}&91.17&2.64&84.21&1.17&\textbf{90.97}&1.04&88.37&\textbf{0.12}&84.16& 24.60 & 85.97 & \textbf{38.3} \\
    \rowcolor{aliceblue} NGF (All layers) & \textbf{1.47} & \textbf{88.65} & 0.42 & \textbf{92.28} & \textbf{2.05} & \textbf{84.61} & \textbf{1.06} & 90.42 & \textbf{0.60} & \textbf{88.74} & 0.18 & \textbf{85.12} & \textbf{19.86} & \textbf{86.30} & 173.2 \\
    
    \bottomrule
    \end{tabular}
    }
    \vspace{-1mm}
    \label{tab:NGF-All}
\end{table*}
 \begin{table*}[t]
    \centering
    \caption{Performance of SGD-Long and NGF while fine-tuning only the last layer of DNN. For SGD-Long, we consider a long purification period with $E_p = 2500$. NGF performance with and without the regularization term underlines the importance of the proposed regularizer. The results shown here are for CIFAR10 dataset.} 
    \scalebox{0.86}{
    \begin{tabular}{c|cc|cc|cc|cc|cc|cc|cc|c}
    \toprule
    \multirow{2}{*}{Methods} & \multicolumn{2}{c|}{Badnets} & \multicolumn{2}{c|}{Blend} & \multicolumn{2}{c|}{Trojan} & \multicolumn{2}{c|}{Dynamic} & \multicolumn{2}{c|}{CLB}& \multicolumn{2}{c|}{SIG} & \multicolumn{2}{c|}{CBA} & Runtime \\ 
    
      &  ASR & ACC & ASR & ACC& ASR & ACC& ASR & ACC&  ASR & ACC& ASR & ACC & ASR & ACC & (Secs.) \\
    \midrule
    Initial&100&92.96&100&94.11&100&89.57&100&92.52&100&92.78&100&88.64 & 93.20 & 90.17& -- \\
    SGD-Long & 82.34 & \textbf{90.68} & 7.13 & \textbf{92.46} & 86.18 & \textbf{87.29} & 57.13 & 90.51 & 13.84 & 88.11 & 0.26 & \textbf{85.74} & 84.41 & \textbf{86.87} & 907.5 \\
    NGF w/o Reg. & 1.91 & 87.65 & \textbf{0.31} & 90.54 & 3.04 & 83.31 & 1.28 & 90.24 & \textbf{0.92} & 87.13 & 0.16 & 84.46 & 25.58 & 84.81 & \textbf{37.8} \\
    \midrule
    \rowcolor{aliceblue} NGF &\textbf{1.86}&88.32&0.38&91.17&\textbf{2.64}&84.21&\textbf{1.17}&\textbf{90.97}&1.04&\textbf{88.37}&\textbf{0.12}&84.16& \textbf{24.60} & 85.97 & 38.3 \\
    
    \bottomrule
    \end{tabular}
    }
    \label{tab:SGD-long}
\end{table*}

\begin{table*}[t]
\centering
\caption{Evaluation of NGF on backdoor attacks with high poison rates, up to 50\%. We consider CIFAR10 dataset and two closely performing defenses for this comparison.
}
\label{tab:poison_rate}
\scalebox{0.85}
{
\begin{tabular}{c|cc|cc|cc|cc|cc|cc|cc|cc|cc}
\toprule
Attack & \multicolumn{6}{c|}{BadNets} & \multicolumn{6}{c|}{Blend} & \multicolumn{6}{c}{Trojan}\\ 
\midrule
Poison Rate&\multicolumn{2}{c|}{25\%}&\multicolumn{2}{c|}{35\%}&\multicolumn{2}{c|}{50\%}&\multicolumn{2}{c|}{25\%}&\multicolumn{2}{c|}{35\%}&\multicolumn{2}{c|}{50\%}&\multicolumn{2}{c|}{25\%}&\multicolumn{2}{c|}{35\%}&\multicolumn{2}{c}{50\%}\\
\midrule
Method &ASR &ACC & ASR &ACC & ASR &ACC&ASR &ACC & ASR &ACC & ASR &ACC&ASR &ACC & ASR &ACC & ASR &ACC \\ \midrule
\textit{No Defense} &100&88.26&100&87.43&100&85.11&100&86.21&100&85.32&100&83.28&100&87.88&100&86.81&100&85.97\\
 ANP & 7.81&82.22&16.35&80.72&29.80&78.27&29.96&\textbf{82.84}&47.02&78.34&86.29&69.15&11.96&76.28&63.99&72.10&89.83&70.02\\
 FT&5.21&78.11&8.39&74.06&11.52&69.81&1.41&68.73&4.56&63.87&7.97&55.70&3.98&76.99&4.71&72.05&5.59&70.98 \\
 \rowcolor{aliceblue} NGF (Ours) & \textbf{2.12}&\textbf{85.50}&\textbf{2.47}&\textbf{84.88}&\textbf{4.53}&\textbf{82.32}&\textbf{0.83}&80.62&\textbf{1.64}&\textbf{79.62}&\textbf{2.21}&\textbf{76.37}&\textbf{3.02}&\textbf{83.10}&\textbf{3.65}&\textbf{81.66}&\textbf{4.66}&\textbf{80.30}\\
 \bottomrule
\end{tabular}}
\end{table*} 
\medskip

\vspace{-2mm}
\subsection{Performance Evaluation of NGF}\label{sec:performance}

In Table~\ref{tab:main}, we present the performance of different defenses for four different datasets. 

\noindent \textbf{CIFAR10:} We consider five \emph{label poisoning attacks}: Badnets, Blend, TrojanNet, Dynamic, and BPPA.  
For TorjanNet, we consider two different variations based on label-mapping criteria: Troj-one and Troj-all. Regardless the complexity of the label-mapping type, our proposed method outperforms all other methods both in terms of ASR and ACC. We also create two variations for Dynamic attack: Dyn-one and Dyn-all. Dynamic attack optimizes for input-aware triggers that are capable of fooling the model; making it more challenging than the static trigger based attacks (Badnets, Blend and Trojan). However, NGF outperforms other methods by a satisfactory margin. We also consider attacks that does not change the label during trigger insertion, \ie \emph{clean label attack}. Two such attacks are CLB and SIG. For further validation of our proposed method, we use \emph{deep feature based attacks}, CBA and FBA. Both of these attacks manipulates deep features for backdoor insertion.
Compared to other defenses, NGF shows better effectiveness against these diverse set of attacks achieving an average drop of $95.01\%$ in ASR while sacrificing an ACC of $3.33\%$ for that. Table~\ref{tab:main} also shows the performance of baseline methods such as I-BAU and AWM. AWM performs similarly as ANP and often struggles to remove the backdoor.

\noindent\textbf{GTSRB:} 
In case of GTSRB, almost all defenses perform similarly for Badnets and Trojan. This, however, does not hold for blend as we achieve an $2.17\%$ ASR improvement over the next best method. The performance is consistent for other attacks as well. Overall, we record an average $97.39\%$ ASR drop with only an $2.79\%$ drop in ACC. \textit{In some cases, ACC for I-BAU are slightly better as it uses a much larger validation size (5\%) for purification than other defense techniques.}



\medskip

\noindent\textbf{ImageNet:} For the scalability test of NGF, we consider two large and widely used datasets, Tiny-ImageNet and ImageNet. In consistence with other datasets, NGF obtains SOTA performance in these diverse datasets too. The effectiveness of ANP reduces significantly for this dataset. In case of large models and datasets, the task of identifying and pruning vulnerable neurons gets more complicated and may result in wrong neurons pruning. \emph{ Note that, we report results for successful attacks only. For attacks such as Dynamic and BPPA (following their implementations), it is challenging to obtain satisfactory attack success rates for Tiny-ImageNet and ImageNet.}   


\begin{table*}[ht]
    \centering
    \caption{ Purification performance for various validation data size. NGF performs well even with very few validation data, \eg 50 data points. All results are for CIFAR10 and Badnets attack.}
    \scalebox{1}{
    \begin{tabular}{c|cc|cc|cc|cc|cc}
    \toprule
         Validation size & \multicolumn{2}{c|}{50}  & \multicolumn{2}{c|}{100} & \multicolumn{2}{c|}{250} & \multicolumn{2}{c|}{350} & \multicolumn{2}{c}{500}   \\
         \midrule
         Method & ASR & CA & ASR & CA & ASR & CA & ASR & CA & ASR & CA \\
         \midrule
         No Defense & 100 & 92.96 & 100 & 92.96 &100 & 92.96 &100 & 92.96  &100 & 92.96 \\
         ANP & 13.66 & 83.99 & 8.35 & 84.47 & 5.72 & 84.70 & 3.78 & 85.26 & 2.84 & 85.96\\
         AWM & 8.51 & 83.63 & 7.38 & 83.71 & 5.16 & 84.52 & 5.14 & 85.80 & 4.34 & 86.17 \\
         \rowcolor{aliceblue} NGF (Ours) & \textbf{6.91} & \textbf{86.82} & \textbf{4.74} & \textbf{86.90} & \textbf{4.61} & \textbf{87.08} & \textbf{2.45} & \textbf{87.74} & \textbf{1.86} & \textbf{88.32} \\
         \bottomrule
    \end{tabular}}
    \label{tab:dif_val}
\end{table*}
\subsection{Ablation Studies}\label{sec:ablation_studies}

\noindent\textbf{Smoothness Analysis of Different Attacks:} We show the relationship between loss surface smoothness and backdoor insertion process in Fig.~\ref{fig:backdoor_eigens}~and~\ref{fig:backdoor_asr}. During backdoor insertion, the model is optimized for two different data distributions: clean and poison.
Compared to a benign model, the loss surface of a backdoor \emph{becomes much sharper as the model becomes well optimized for both distributions}, \ie model has both high ASR and high ACC. Backdoor and benign models are far from being well-optimized at the beginning of training. The difference between these models is prominent once the model reaches closer to the final optimization point. As shown in Fig.~\ref{fig:backdoor_asr}, the training becomes reasonably stable after 100 epochs with ASR and ACC near saturation level. Comparing $\lambda_\mathsf{max}$ of benign and all backdoor models after 100 epochs, we notice a sharp contrast  in Fig.~\ref{fig:backdoor_eigens}. This validates our previous claim on loss surface smoothness of benign and backdoor models. 
During the purification period, as shown in Fig.~\ref{fig:purified_eigens} and~\ref{fig:purified_asr}, the model is optimized to a smoother minima. As a result, ASR becomes close to 0 while retaining good clean test performance. Note that, we calculate loss Hessian and $\lambda_\mathsf{max}$ using all DNN parameters. This indicates that changing the parameters of only one layer impacts the loss landscape of the whole network. Even though the CNN-backbone parameters are frozen, NGF changes the last layer in a way such that the whole backdoor network behaves differently, \ie like a benign model.       
\medskip 

\noindent \textbf{Evaluation of Different Optimizers:} We compare the performance of NGF with different variants of first-order optimizers: 
(i) \emph{AdaGrad}~\citep{duchi2011adaptive}, 
(ii) \emph{RMSProp}~\citep{hinton2012neural}, 
(iii) \emph{Adam}~\citep{kingma2014adam}, and
(iv) Sharpness-Aware Minimization (\emph{SAM})~\citep{foret2021sharpnessaware} is a recently proposed SGD-based optimizer that explicitly penalizes the abrupt changes of loss surface by bounding the search space within a small region. This forces the changes of model parameters so that the optimization achieves a smoother loss surface. 
Table~\ref{tab:different_optimizer} shows that NGF outperforms all of these variants of first-order optimizer by a huge margin. At the same time, the proposed method achieves comparable clean test performance. 
Although SAM usually performs better than vanilla SGD in terms of smooth DNN optimization, SAM’s performance in shallow network scenarios (i.e., our case) is almost similar to vanilla SGD. Two potential reasons behind this poor performance are (i) using a predefined local area to search for maximum loss, and (ii) using the `Euclidean distance' metric instead of the geometric distance metric. In contrast, NGD with curvature geometry aware Fisher Information Matrix can successfully avoid such bad minima and optimizes to global minima.

\noindent \textbf{Runtime Analysis:} In Table~\ref{tab:runtime_dataset}, we show the average runtime for different defenses. Similar to purification performance, purification time is also an important indicator to measure the success of a defense technique. In Section~\ref{sec:performance}, we already show that our method outperforms other defenses in most of the settings. As for the run time, our method completes the purification (for CIFAR10) in just $38.3$ seconds; which is almost half as compared to FT. The time advantage of our method also holds for large datasets and models, \eg ImageNet and ResNet50. Runtime comparison with other defenses is in the \emph{supplementary material}.  

\medskip

\noindent \textbf{Fine-tuning All Layers:}
We have considered fine-tuning all layers fusing NGF and SGD. Note that vanilla FT does fine-tune all layers. We report the performance of NGF for all layers in Table~\ref{tab:NGF-All}. While fine-tuning all layers seems to improve the performance, it takes almost 6$\times$ more computational time than NGF on the last layer. We also show the results of SAM and SGD while fine-tuning all layers: we term them as vanilla FT (SAM) and vanilla FT (SGD). SAM has a slightly better ASR performance compared to SGD, which aligns with our smoothness hypothesis as SAM usually leads to smoother loss surface.  As for execution time, each SAM update requires 2 backpropagation operations while non-SAM update (SGD, Adam, etc.) requires only \textbf{1} backpropagation. This makes vanilla FT (SAM) slower than vanilla FT (SGD) which is not desirable for backdoor purification techniques.

\medskip

              






\noindent \textbf{Effect of Proposed Regularizer:} In this section, we analyze the effect of regularizer and long training with SGD. The effect of our clean distribution-aware regularizer can be observed in Table~\ref{tab:SGD-long}. NGF with the proposed regularizer achieves a 1\% clean test performance improvement over vanilla NGF. For long training with SGD (SGD-Long), we fine-tune the last layer for 2500 epochs. Table~\ref{tab:SGD-long} shows the evaluations of SGD-Long on 7 different attacks. Even though the ASR performance improves significantly for CLB and SIG attacks, SGD-based FT still severely underperforms for other attacks. Moreover, the computational time increases significantly over NGF. Thus, our choice of \emph{NGD-based FT as a fast and effective backdoor purification technique} is well justified.  

\medskip

\noindent \textbf{Effect of Clean Validation Data Size:}
We also present how the total number of clean validation data can impact the purification performance. In Table~\ref{tab:dif_val}, we see the change in performance while gradually reducing the validation size from 1\% to 0.1\%. We consider Badnets attack on the CIFAR10 dataset for this evaluation. Even with only 50 (0.1\%) data points, NGF can successfully remove the backdoor by bringing down the attack success rate (ASR) to 6.91\%. We also consider AWM performance for this comparison. For both ANP and AWM, reducing the validation size has a severe impact on test accuracy (ACC). 

\medskip

\noindent \textbf{Strong Backdoor Attacks:} By increasing the poison rates, we create stronger versions of different attacks against which most defense techniques fail quite often. We use 3 different poison rates, $\{25\%, 35\%,50\%\}$. 
We show in Table~\ref{tab:poison_rate} that NGF is capable of defending very well even with a poison rate of $50\%$, achieving a significant ASR improvement over FT. 
Furthermore, there is a sharp difference in classification accuracy between NGF and other defenses. For $25\%$ Blend attack, however, ANP offers a slightly better performance than our method. However, ANP performs poorly in removing the backdoor as it obtains an ASR of $29.96\%$ compared to $0.83\%$ for NGF. 

\section{Discussion}

\noindent{\bf Why Smoothness is Key to Removing Backdoor?}\label{sec:smooth_backdoor_relationship}
One key observation from the smoothness study is that: there exists a key difference between weight-loss surface smoothness (estimated by \textit{loss hessian}) of a backdoor and a benign model w.r.t. clean distribution—the weight-loss surface of a backdoor model is less smooth compared to a benign model. To further elaborate, let us consider feeding a clean sample to a backdoor model. By definition, it will predict the correct ground truth label. Now, consider feeding a sample with a backdoor trigger on it. The model will predict the adversary-set target label implying significant changes in prediction distribution. This significant change can be explained by the surface smoothness. In order to accommodate this significant change in prediction, the model must adjust itself accordingly. Such adjustment leads to non-smoothness in the weight-loss surface. \textit{A non-smooth surface causes significant changes in loss gradient for specific inputs.}  In our case, these specific inputs are backdoor-triggered samples.  As the magnitude of a trigger is usually very small compared to the total input magnitude, the model has to experience quite a significant change in its weight space to cause large loss changes. We characterize this change in terms of smoothness. As for backdoor removal, we claim that making the non-smooth weight loss surface smoother removes the backdoor behavior. Based on the above discussion, a smoother surface should not cause a large change in loss or model predictions corresponding to backdoor-related perturbations or triggers. In summary, for a model to show certain backdoor behavior, there are some specific changes that take place in the weight space. In this work, we try to explain these changes regarding weight-loss surface smoothness.  Our comprehensive empirical evaluations support our intuition well. 

\medskip

\noindent{\bf Why the Classification Layer?}
We further offer an explanation as to why we choose to fine-tune the classification layer instead of any other layer, e.g. input layer. The classification layer is mostly responsible
for the final prediction in a DNN. Depending on the extracted
features by the CNN backbone, the classifier learns the decision
boundary between these features and renders a prediction.
While backdooring we change the input features slightly
(by inserting triggers) so that the classifier makes a wrong
prediction. If we can make the classifier invariant to these
slight input changes, the effect of the backdoor should be
removed. Thus, compared to other layers of the DNN, the
classifier plays a more important role in the overall backdoor
insertion and removal process. Another reason for fine-tuning
only the last layer is for better computational efficiency; which
is one of the most important aspects of a backdoor defense
technique.

\medskip

\noindent{\bf Why Different Metrics for Smoothness and NGF?}
It is natural to probe that the same technique---either the Hessian of loss or the Fisher Matrix---could be used for both our smoothness analysis and the development of the proposed method. However, our particular choice is driven by the trade-off related to these two matrics---\emph{computational efficiency and performance}. Note that computing Hessian is more expensive than the FIM. On the other hand, Hessian is a slightly better indicator of smoothness due to its superior effectiveness in capturing loss-surface geometry. Therefore, we can choose either one of the metrics. Since smoothness analysis is performed in an offline manner and only once (for each instance), we choose the better performing one, i.e. Hessian of loss, for smoothness analysis. As to why we choose the Fisher matrix in developing our proposed method, we need to design a runtime-efficient method with good performance. Since we have to calculate either FIM or Hessian in each iteration of the update, it becomes harder to choose Hessian over FIM for the development of the proposed method. Given the potential of a higher trade-off value of the Fisher-information matrix, we develop our method based on it.

\medskip

\noindent{\bf Why Fine-tuning Negatively Impacts ACC?}
It is observable that no matter which defense techniques we use the clean test accuracy consistently drops for all datasets. We offer an explanation for fine-tuning based techniques as NGF is one of them. As we use a small validation set for fine-tuning, it does not necessarily cover the whole training data distribution. Therefore, fine-tuning with this small amount of data bears the risk of overfitting and reduced clean test accuracy. This is more prominent when we fine-tune all layers of the network (vanilla FT in Table 2). Whereas, NGF fine-tunes only 1 layer which shows to be better in terms of preserving clean test accuracy. 

\section{Conclusion}
We propose a novel backdoor purification technique based on natural gradient descent fine-tuning. The proposed method is motivated by our analysis of loss surface smoothness and its strong correlation with the backdoor insertion and purification processes. As a backdoor model has to learn an additional data distribution, it tends to be optimized to bad local minima or sharper minima compared to a benign model. We argue that the backdoor can be removed by re-optimizing the model to a smoother minima. We further argue that fine-tuning a single layer is enough to remove the backdoor. Therefore, in order to achieve a smooth minima in a single-layer fine-tuning scenario, we propose using an FIM-based DNN objective function and minimizing it using a curvature-aware NGD optimizer. Our proposed method achieves SOTA performance in a wide range of benchmarks. Since we fine-tune only one layer the training time overhead reduces significantly, making our method one of the fastest among SOTA defenses. 

\noindent{\bf Limitations and future works. }
Our extensive empirical studies on loss surface smoothness show its relationship with backdoor insertion and removal. However, we left the mathematical analysis of this relationship for future studies. Such analysis should be able to address the nature of convergence under different purification settings, e.g., the number of validation samples, number of iterations, number of fine-tuning layers, etc. Although we verify the smoothness hypothesis empirically, the mathematical analysis will give us more insight into understanding the backdooring process.  Although we only experimented with CNN-based architectures, our findings should also hold for attention-based vision transformer (ViT)~\cite{dosovitskiy2020image} architecture. Nevertheless, further study is required to verify the smoothness claims for the ViT architecture. It is known that the attention mechanism and residual connection generally lead the optimization towards smooth minima. However, how the backdooring process interferes with this optimization must be explored properly. In future, we aim to extend our smoothness analysis to 3D point-cloud attacks as well as contrastive backdoor attacks.
\bibliographystyle{IEEEtran}
\bibliography{ref}

\begin{IEEEbiography}[{\includegraphics[width=1in,height=1in,clip,keepaspectratio]{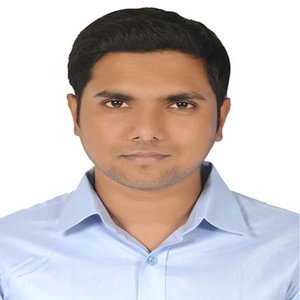}}]{Nazmul Karim}
received his B.S. degree in electrical engineering from the Bangladesh University of Engineering and Technology, Dhaka, Bangladesh, in 2016. He is currently working toward the Ph.D. degree in electrical engineering at the University of Central Florida. His current research interests lie in the areas of machine learning, signal processing, and linear algebra. Nazmul's awards and honors include the University of Central Florida Multidisciplinary Doctoral Fellowship and Bangladesh University of Engineering and Technology Merit Scholarship.
\end{IEEEbiography}


\vspace{-1cm}
\begin{IEEEbiography}[{\includegraphics[width=1in,height=1.25in,clip,keepaspectratio]{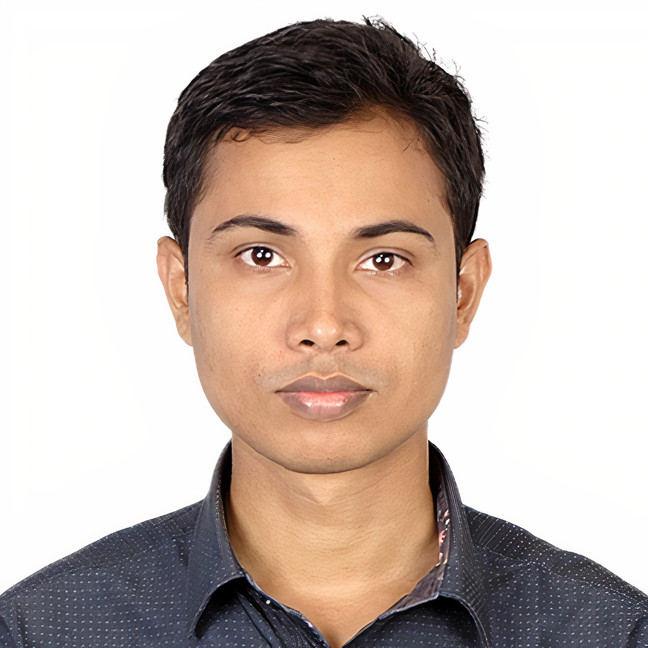}}]{Abdullah Al Arafat}
is currently pursuing his Ph.D. in Computer Science at the North Carolina State University, Raleigh, North Carolina, USA. He received his BS in Electrical Engineering from Bangladesh University of Engineering and Technology (BUET), Bangladesh in 2016 and MS in Computer Engineering from the University of Central Florida, Orlando, Florida, USA in 2020. His research interests include Scheduling Theory, Algorithms, and Real-Time \& Intelligent Systems. 
 
\end{IEEEbiography}

\vspace{-1cm}
\begin{IEEEbiography}[{\includegraphics[width=1in,height=1.25in,clip,keepaspectratio]{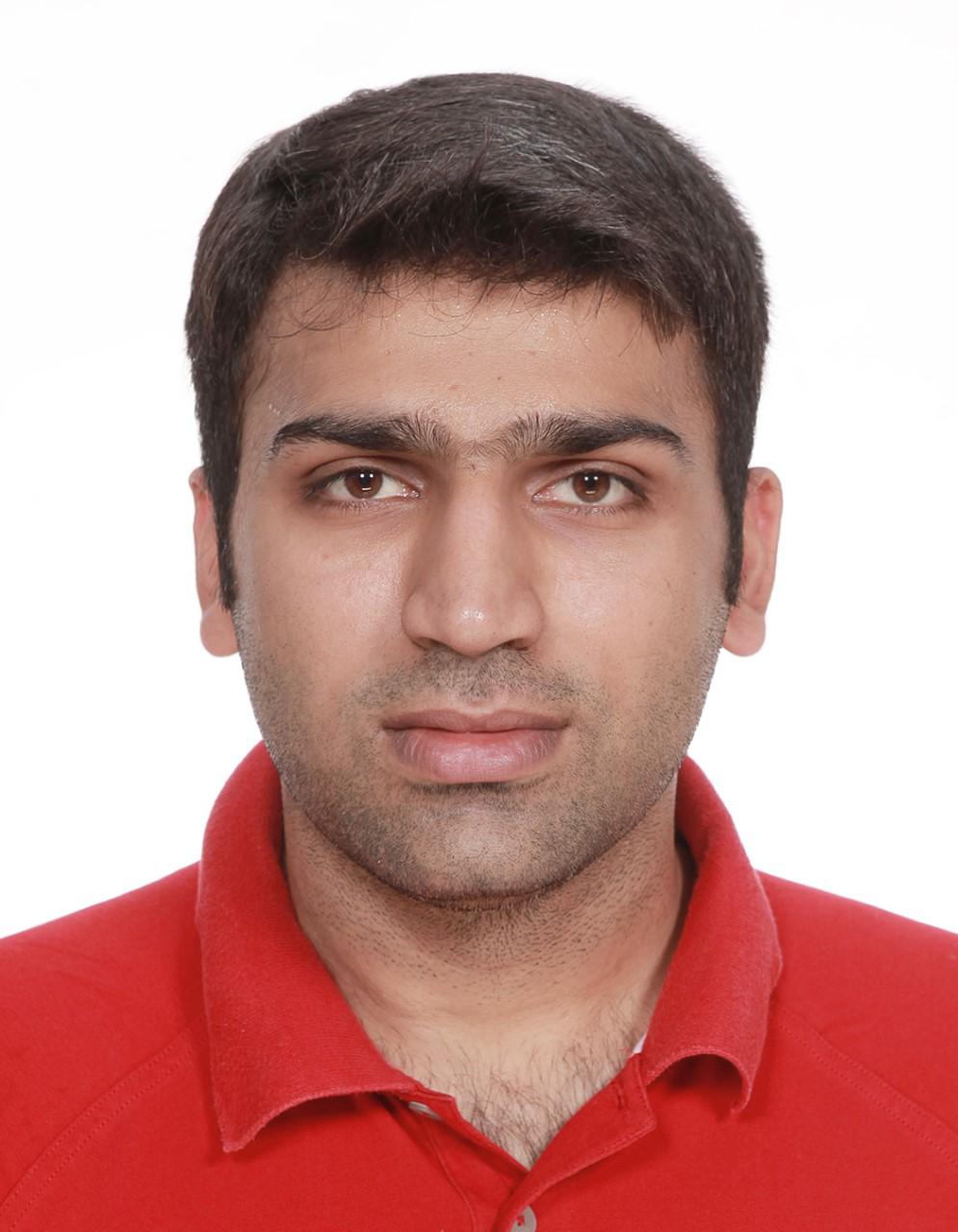}}]{Umar Khalid}
is now a Ph.D. candidate at center of research in Computer Vision, University of Central Florida, USA. He received his master’s degree from the prestigious Shanghai Jiao Tong University and his bachelor’s degree from the National University of Science and Technology of Pakistan. His research interests include AI security, Federated Learning, and video understanding.
 
\end{IEEEbiography}

\vspace{-1cm}
\begin{IEEEbiography}[{\includegraphics[width=1in,height=1.25in,clip,keepaspectratio]{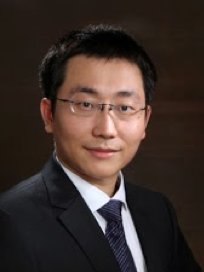}}]{Zhishan Guo}
(Senior Member, IEEE) is an Associate Professor with the Department of Computer Science, North Carolina State University, Raleigh, NC, USA. He received the B.Eng. degree (with honor) in computer science and technology from Tsinghua University, Beijing, China, in 2009, the M.Phil. degree in mechanical and automation engineering from The Chinese University of Hong Kong, Hong Kong, in 2011, and the Ph.D. degree in computer science from the University of North Carolina at Chapel Hill, Chapel Hill, NC, USA, in 2016. 
His current research interests include real-time and cyber-physical systems, neural networks, and computational intelligence. He has received best paper awards from flagship conferences such as RTSS and EMSOFT.
 
\end{IEEEbiography}

\vspace{-0.75 cm}
\begin{IEEEbiography}[{\includegraphics[width=1in,height=1.25in,clip,keepaspectratio]{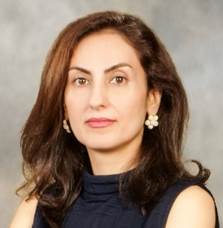}}]{Nazanin Rahnavard}
 (S’97-M’10, SM’19) received her Ph.D. in the School of Electrical and Computer Engineering at the Georgia Institute of Technology, Atlanta, in 2007. She is a Professor in the Department of Electrical and Computer Engineering at the University of Central Florida, Orlando, Florida. Dr. Rahnavard is the recipient of NSF CAREER award in 2011. She has interest and expertise in a variety of research topics in deep learning, communications, networking, and signal processing areas. She serves on the editorial board of the Elsevier Journal on Computer Networks (COMNET) and on the Technical Program Committee of several prestigious international conferences.
\end{IEEEbiography}
\end{document}

%% file: math_commands.tex
\usepackage{amsmath,amsfonts,bm}

\def\eqref#1{equation~\ref{#1}}

\def\1{\bm{1}}

\DeclareMathAlphabet{\mathsfit}{\encodingdefault}{\sfdefault}{m}{sl}
\SetMathAlphabet{\mathsfit}{bold}{\encodingdefault}{\sfdefault}{bx}{n}

\DeclareMathOperator*{\argmax}{arg\,max}
\DeclareMathOperator*{\argmin}{arg\,min}

%% file: intro_fisher.tex
\section{Introduction}\label{sec:intro}
Training a deep neural network (DNN) with a fraction of poisoned or malicious data is often security-critical since the model can successfully learn both clean and adversarial tasks equally well.
This is prominent in scenarios where one outsources the DNN training to a vendor. In such scenarios, an adversary can mount backdoor attacks~\citep{gu2019badnets,chen2017targeted} through poisoning a portion of training samples so that the model will misclassify any sample with a \emph{particular trigger} or \emph{pattern} to an adversary-set label. Whenever a DNN is trained in such a manner, it becomes crucial to remove the effect of backdoor before deploying it for a real-world application.    

Different defense techniques~\citep{liu2018fine,wang2019neural,wu2021adversarial, li2021anti,zheng2022data} have been proposed for purifying backdoor. 
Techniques such as fine-pruning~\citep{liu2018fine} and adversarial neural pruning~\citep{wu2021adversarial} require a long training time due to iterative searching criteria. Furthermore, the purification performance deteriorates significantly as the attacks get stronger.
In this work, we explore the backdoor insertion and removal phenomena from the DNN optimization point of view. Unlike a benign model, a backdoor model is forced to learn two different data distributions: clean data distribution and poisoned/trigger data 
distribution. Having to learn both distributions,
backdoor model optimization usually leads to a \emph{bad local minima} or sharper minima \emph{w.r.t.} clean distribution. 
We claim that backdoor can be removed by re-optimizing the model to a smoother minima. One easy re-optimization scheme could be simple DNN weights fine-tuning with a few clean validation samples. However, fine-tuning all DNN parameters often requires huge computational cost and may result in sub-par clean test performance after purification. Therefore, we intend to \emph{fine-tune only one layer} to effectively remove the backdoor. 

Fine-tuning only one layer creates a shallow network scenario where SGD-based optimization becomes a bit challenging. \cite{choromanska2015loss} claims that the probability of finding bad local minima or poor quality solution increases as the network size decreases. Even though there are good-quality solutions, it usually requires exponentially long time to find those minima~\citep{choromanska2015loss}. As a remedy to this, we opt to use a curvature aware optimizer, Natural Gradient Decent (NGD), that has \emph{higher probability of escaping the bad local minima as well as faster convergence rate}, specifically in the shallow network scenario \citep{amari1998natural,martens2015optimizing}. To this end, we propose a novel backdoor purification technique---\underline{N}atural \underline{G}radient \underline{F}ine-tuning (NGF)---which focuses on removing backdoor through fine-tuning \emph{only one layer}. However, straightforward application of NGF with simple cross-entropy (CE) loss may result in poor clean test performance. To boost this performance, we use a {clean distribution-aware} regularizer that prioritizes the update of parameters sensitive to clean data distribution. Our proposed method achieves SOTA performance in a wide range of benchmarks, \eg four different datasets including \emph{ImageNet}, 13 recent backdoor attacks \etc Our contributions can be summarized as follows:
\begin{itemize}
    \vspace{-1mm}
    \item We analyze the loss surface characteristics of a DNN during backdoor insertion and purification processes. Our analysis shows that the optimization of a backdoor model leads to a \emph{bad local minima} or sharper minima compared to a benign model. We argue that backdoor can be purified by re-optimizing the model to a smoother minima and simple fine-tuning can be a viable way for that. 
    To the best of our knowledge, this is the first work that studies the correlation between loss-surface smoothness and backdoor purification.
    
    \item We conduct additional studies on backdoor purification process while fine-tuning different parts of a DNN. We observe that SGD-based one-layer fine-tuning fails to escape bad local minima and a loss surface geometry-aware optimizer can be an easy fix to this. 
    
    
     
    \item We propose a novel backdoor purification technique based on Natural Gradient Fine-tuning (NGF).  In addition, we employ a {clean distribution-aware} regularizer to boost the clean test performance of our proposed method. NGF outperforms recent SOTA methods in a wide range of benchmarks.   
    
    
    
\end{itemize}

%% file: related_work_fisher.tex
\section{Related Work}
This section discusses the related works related to the backdoor attack methods and the defenses for backdoor attacks. 

\noindent\textbf{Backdoor Attacks.} 
Backdoor attacks in deep learning models aim to manipulate the model to predict adversary-defined target labels in the presence of backdoor triggers in input while the model predicts true labels for benign input~\cite{guo2022overview}. Monoj \etal~\cite{manoj2021excess} formally analyzed DNN and revealed the intrinsic capability of DNN to learn backdoors. 
Backdoor triggers can exist in the form of dynamic patterns~\cite{li2020backdoor}, 
a single pixel~\cite{tran2018spectral}, 
sinusoidal strips~\cite{barni2019new}, human imperceptible noise~\cite{zhong2020backdoor}, natural reflection~\cite{liu2020reflection}, adversarial patterns~\cite{zhang2021advdoor}, blending backgrounds~\cite{chen2017targeted}, hidden trigger~\cite{saha2020hidden}~\etc
Based on target labels, existing backdoor attacks can generally be classified as poison-label or clean-label backdoor attacks.
In poison-label backdoor attack, the target label of the poisoned sample is different from its ground-truth label, \eg BadNets~\cite{gu2019badnets}, Blended attack~\cite{chen2017targeted}, SIG attack~\cite{barni2019new}, WaNet~\cite{nguyen2021wanet}, Trojan attack~\cite{liu2017trojaning}, and BPPA~\cite{wang2022bppattack}. 
 Contrary to the poison-label attack, clean-label backdoor attack doesn't change the label of the poisoned sample~\cite{turner2018clean, huang2022backdoor,zhao2020clean}. 
Recently,~\cite{saha2022backdoor} studied backdoor attacks on self-supervised learning. All these attacks emphasized the severity of backdoor attacks and the necessity of efficient removal/purification methods. 

\noindent\textbf{Backdoor Defenses.} 
Existing backdoor defense methods can be categorized into backdoor detection or purifying techniques. Detection based defenses include trigger synthesis approach~\cite{wang2019neural,qiao2019defending,guo2020towards,shen2021backdoor,dong2021black,guo2021aeva,xiang2022post,tao2022better}, or malicious samples filtering based techniques~\cite{tran2018spectral, gao2019strip,chen2019deepinspect}. However, these methods only detect the existence of backdoor without removing it. 
Backdoor purification defenses can be further classified as training time defenses and inference time defenses. Training time defenses include model reconstruction approach~\cite{zhao2020bridging,li2021neural}, poison suppression approach ~\cite{hong2020effectiveness,du2019robust,borgnia2021strong}, and pre-processing approaches~\cite{li2021anti,doan2020februus}. Although training time defenses are often successful, they suffer from huge computational burden and less practical considering attacks during DNN outsourcing. Inference time defenses are mostly based on pruning approaches such as~\cite{ koh2017understanding, ma2019nic, tran2018spectral,diakonikolas2019sever,steinhardt2017certified}. 
Pruning-based approaches are typically based on model vulnerabilities to backdoor attacks. For example, MCR~\cite{zhao2020bridging} and CLP~\cite{zheng2022data} analyzed node connectivity and channel Lipschitz constant to detect backdoor vulnerable neurons. ANP~\cite{wu2021adversarial} prune neurons through backdoor sensitivity analysis using adversarial search on the parameter space. Instead, we propose a simple one-layer fine-tuning based defense that is both fast and highly effective. To remove backdoor, our proposed method revisits the DNN fine-tuning paradigm from a novel point of view---the relation between backdoor training and loss surface geometry (please refer to Sec.~\ref{sec:smoothAnalysis} for details)---allowing us to fine-tune only one-layer. 

%% file: analysis_fisher.tex
\begin{table*}[t]
    \centering
    \caption{ Backdoor removal performance when we fine-tune only the classifier (Cls.), only the CNN backbone (CNN-Bbone), or the full network (Full-Net). Fine-tuning only the last layer creates a shallow network scenario. In such a scenario, there is a high probability that SGD does not escape bad local minima. Whereas, NGF consistently  optimizes to a smooth minima (indicated by low $\lambda_\mathsf{max}$ for 6 different attacks), resulting in backdoor removal, \ie low ASR and high ACC. We consider CIFAR10 dataset and PreActResNet18 architecture for all evaluations. A clean validation set is used for all purification.} 
    \scalebox{0.9}{
    \begin{tabular}{c|cccc|cccc|cccc|cccc}
    \toprule
     FT & \multicolumn{4}{c|}{Badnets} & \multicolumn{4}{c|}{Blend} & \multicolumn{4}{c|}{Trojan} & \multicolumn{4}{c}{Dynamic} 
     \\ 
    
     Methods & $\lambda_\mathsf{max}$ &  Tr(H) & ASR & ACC & $\lambda_\mathsf{max}$& Tr(H) & ASR & ACC & $\lambda_\mathsf{max}$ & Tr(H) & ASR & ACC& $\lambda_\mathsf{max}$& Tr(H) & ASR & ACC
     \\
    \midrule
    Initial&573.8& 6625.8 & 100&92.96&715.5& 7598.3& 100&94.11&616.3& 8046.4& 100&89.57&564.2& 7108.5& 100&92.52
    \\
    Full-Net. & 4.42 & 25.36 & 4.87 & 85.92 & 4.65 & 27.83 &  4.77 & 87.61 & 3.41 & 26.15 & 3.78 & 82.18 & 2.34 & 15.82 & 4.73& 88.61  
    \\
    CNN-Bbone. & 4.71 & 28.08 & 5.03 & 85.64 & 5.14 & 31.16 & 4.92 & 87.24 & 4.19 & 29.67 & 3.95 & 81.86 & 2.46 & 16.08 & 5.11 & 87.54 
    \\ 
    Cls. (SGD) & 556.1 & 6726.3 & 98.27 & \textbf{90.17} & 541.7 & 5872.5 & 97.29 & \textbf{93.48} & 613.0 & 6829.7 & 96.25 & \textbf{87.36} & 446.5& 5176.6 & 93.58&\textbf{91.36}
    \\
    \midrule
     \rowcolor{aliceblue} Cls. (NGF)&\textbf{2.79}&\textbf{16.94}&\textbf{1.86}&88.32&\textbf{2.43}&\textbf{16.18}&\textbf{0.38}&91.17&\textbf{2.74}&\textbf{17.32}&\textbf{2.64}&84.21&\textbf{1.19}& \textbf{8.36} &\textbf{1.17}&90.97
     \\
    \bottomrule
    \end{tabular}
    }
    \vspace{-2.5mm}
    \label{tab:eigen-ASR-ACC}
\end{table*}